\documentclass[preprint,12pt]{elsarticle}
\usepackage{amsfonts}  
\usepackage{amsmath}
\usepackage{xcolor}
\usepackage{amsthm}
\usepackage{multirow}
\usepackage{booktabs}
\usepackage{graphicx}
\usepackage{subcaption}
\usepackage{enumitem}
\usepackage{float}
\usepackage{algorithm}
\usepackage{algorithmic}
\usepackage{url} 
\usepackage{longtable}
\usepackage{multirow}
\usepackage{caption}
\usepackage{hyperref}
\usepackage{bm}
\usepackage{booktabs}

\begin{document}

\begin{frontmatter}

\title{SeqBattNet: A Discrete-State Physics-Informed Neural Network with Aging Adaptation for Battery Modeling}

\author[1]{Khoa~Tran\fnref{equal}}
\ead{khoa.tran@aiware.website}

\author[2]{Hung-Cuong Trinh}
\ead{trinhhungcuong@tdtu.edu.vn}

\author[3]{Vy-Rin~Nguyen}
\ead{RinNV@fe.edu.vn}

\author[4,5]{T. Nguyen-Thoi}
\ead{trung.nguyenthoi@vlu.edu.vn}

\author[6,7]{Vin Nguyen-Thai\corref{cor1}}
\ead{vin.nguyenthai@vlu.edu.vn}

\cortext[cor1]{Corresponding author: \href{mailto:vin.nguyenthai@vlu.edu.vn}{vin.nguyenthai@vlu.edu.vn}}

\affiliation[1]{organization={AIWARE Limited Company},  
addressline={17 Huynh Man Dat Street, Hoa Cuong Bac Ward, Hai Chau District},  
city={Da Nang},  
postcode={550000},  
country={Vietnam}}  
 
\affiliation[2]{organization={Faculty of Information Technology, Ton Duc Thang University},  
city={Ho Chi Minh City},  
postcode={700000},  
country={Vietnam}}  

\affiliation[3]{organization={Software Engineering Department, FPT University},  
city={Da Nang},  
postcode={550000},  
country={Vietnam}} 

\affiliation[4]{organization={Laboratory for Applied and Industrial Mathematics, Institute for Computational Science and Artificial Intelligence, Van Lang University},  
city={Ho Chi Minh City},  
postcode={700000},  
country={Vietnam}}  

\affiliation[5]{organization={Faculty of Mechanical-Electrical and Computer Engineering, School of Technology, Van Lang University},  
city={Ho Chi Minh City},  
postcode={700000},  
country={Vietnam}}  

\affiliation[6]{organization={Laboratory for Computational Mechanics, Institute for Computational Science and Artificial Intelligence, Van Lang University},  
city={Ho Chi Minh City},  
postcode={700000},  
country={Vietnam}}  

\affiliation[7]{organization={Faculty of Civil Engineering, School of Technology, Van Lang University},  
city={Ho Chi Minh City},  
postcode={700000},  
country={Vietnam}}

\begin{abstract}
Accurate battery modeling is essential for reliable state estimation in modern applications, such as predicting the remaining discharge time and remaining discharge energy in battery management systems. Existing approaches face several limitations: model-based methods require a large number of parameters; data-driven methods rely heavily on labeled datasets; and current physics-informed neural networks (PINNs) often lack aging adaptation, or still depend on many parameters, or continuously regenerate states. In this work, we propose \textbf{SeqBattNet}, a discrete-state PINN with built-in aging adaptation for battery modeling, to predict terminal voltage during the discharge process. SeqBattNet consists of two components: (i) an encoder, implemented as the proposed HRM-GRU deep learning module, which generates cycle-specific aging adaptation parameters; and (ii) a decoder, based on the equivalent circuit model (ECM) combined with deep learning, which uses these parameters together with the input current to predict voltage. The model requires only three basic battery parameters and, when trained on data from a single cell, still achieves robust performance. Extensive evaluations across three benchmark datasets (TRI, RT-Batt, and NASA) demonstrate that SeqBattNet significantly outperforms classical sequence models and PINN baselines, achieving consistently lower RMSE while maintaining computational efficiency. 

\textit{Keywords}---Battery Management Systems, Lithium-ion Batteries, Remaining Discharge Time, Battery Modeling.
\end{abstract}

\end{frontmatter}

\section{Introduction}
In recent years, the use of battery-powered devices—such as electric vehicles, portable electronic devices, and unmanned aerial vehicles—has rapidly increased, highlighting the critical role of batteries in modern technology~\cite{wang2020framework}. Among various battery types, lithium-ion batteries (LIBs) are the most widely used due to their high energy density and long cycle life. The global LIB market is projected to exceed 170 billion USD by 2030~\cite{ma2023two}. However, LIBs exhibit strong coupling effects and nonlinear behavior, posing significant challenges for their control and management. 

Therefore, an accurate Battery Management System (BMS)~\cite{gabbar2021review} is essential for monitoring the health, safety, and performance of batteries. A typical BMS encompasses two core functions: life cycle prognosis and in-cycle state estimation. Life cycle prognosis focuses on long-term performance metrics such as state of health (SOH)\cite{qu2019neural}, remaining useful life (RUL)\cite{liang2024hybrid, liu2024hybrid, ma2024accurate}, and cycle life prediction~\cite{severson2019data, ma2023two}. These aspects have attracted considerable research interest due to their critical role in extending battery lifespan and ensuring system reliability. On the other hand, in-cycle state estimation involves real-time indicators such as state of charge (SOC)\cite{liu2020novel, wang2025adaptive}, remaining discharge time (RDT)\cite{wang2020framework, chen2019particle, quinones2018remaining}, and remaining discharge energy (RDE)~\cite{lai2022remaining, tu2024remaining, quinones2018remaining}, to enhance user experience (range and time-to-empty guidance, reduced range anxiety, charge scheduling)~\cite{noel2019fear}. 

In approaches to RDT prediction and RDE estimation, battery modeling~\cite{wang2023inherently} is a crucial component for predicting terminal voltage from the input current, in order to determine the time at which the predicted voltage falls below the cut-off voltage (the end-of-discharge time), which is then used to estimate the remaining discharge time and, subsequently, the remaining discharge energy. Battery modeling has attracted increasing attention in recent years, particularly in RDT~\cite{chen2019particle, wang2023inherently, biggio2023ageing} and RDE estimation~\cite{hatherall2023remaining, tu2024remaining}. This is generally classified into three categories: model-based, data-driven, and physics-informed neural network (PINN) approaches.

Model-based methods characterize battery dynamics and degradation via physic-based formulations, including electrochemical models~\cite{gu2000thermal}, and ECMs~\cite{lai2018comparative}. ECMs approximate a cell with an open-circuit voltage (OCV) source in series with an ohmic resistance and one or more RC polarization branches~\cite{liaw2004modeling}. Within electrochemical approaches, the pseudo two-dimensional (P2D/DFN)~\cite{csomos2023comparison} and single-particle model (SPM)~\cite{cen2020lithium} are prevalent. Although high-fidelity, these models demand many difficult-to-identify parameters and careful calibration, rendering parameter identification expensive and real-time deployment computationally burdensome~\cite{ali2024comparison}.

Data-driven methods use machine learning models to learn battery behavior directly from measured data rather than from explicit physical equations. Consequently, they require large, well-labeled datasets spanning diverse operating conditions to capture variability in behavior. \cite{biggio2023ageing} proposed \textit{Dynaformer}, which employs an encoder to generate context for the current battery degradation state and passes this context to a decoder that, together with the input current, predicts the terminal voltage. In general, data-driven methods are less commonly used in battery modeling for RDT or RDE estimation, because reliable predictions require large, diverse labeled datasets to generalize across operating conditions. However, generating battery labels is costly and time-consuming~\cite{tan2025batterylife}, and data-driven methods lack interpretability, making them less practical for some real-world applications.

PINN methods address the limitations of model-based and data-driven approaches, requiring only a few physical parameters and a relatively small amount of labeled data, while still achieving high prediction performance. \cite{tu2024remaining} uses an equivalent circuit model (ECM) called the nonlinear double capacitor (NDC) model. The NDC model generates states ($V_b$, $V_s$, $V_1$, $T_{\mathrm{core}}$, $T_{\mathrm{surf}}$) in derivative form, which are then propagated using ordinary differential equations (ODEs). The generated states, together with the input current, are fed into a feedforward neural network to predict the terminal voltage. However, this method relies on continuous ODEs to generate states based on preset start and end times. If the end time is set too high, the predicted terminal voltage can fall far below the cutoff voltage, causing the remaining operation time to be overestimated, which is not practical for RDT and RDE prediction. \cite{wang2023inherently} proposed a discrete-state PINN that overcome this limitation of continuous modeling by updating the state at each time step, allowing the prediction to stop proactively once the predicted terminal voltage falls below the cutoff voltage. The approach enhances the traditional ECM by using feed-forward networks (FNNs) to predict SOC, OCV, and $1/R_{sp}$, and proposes a loss function that achieves high prediction accuracy with only 30 training samples. However, this model still requires physical parameters such as $R_p$, $R_s$, $C_{sp}$, and $C_s$, which must be obtained through measurement or grid search. In addition, it does not incorporate aging adaptation, leading to biased predictions under degraded states of health. The paper's data-splitting strategy also introduces data leakage, since in every discharge cycle the first 60 data points are used for training while the remaining points are used for testing. As both training and testing data come from the same cell, this reduces the generalizability of the evaluation. To overcome the aforementioned limitation, we propose the following contributions:
\begin{itemize}    
    \item We propose \textbf{SeqBattNet}, a discrete-state PINN for battery modeling that addresses the limitations of: (i) model-based methods, which require hard-to-measure physical parameters; (ii) data-driven methods, which demand large datasets; and (iii) prior PINN baselines, which rely on continuous ODEs, lack proactive stopping capability, or require additional hard-to-determine physical parameters such as $R_p$, $R_s$, $C_{sp}$, and $C_s$ beyond the available parameters $V_0$, $V_{\mathrm{EOD}}$, and $C$, or do not incorporate aging adaptation. Unlike these approaches, SeqBattNet introduces a discrete-state formulation that enables proactive termination once the terminal voltage falls below the cutoff voltage. Moreover, it requires minimal physical knowledge---specifically, $V_0$, $V_{\mathrm{EOD}}$, and $C$, which are readily available from the battery cell---and can be trained on data from a single cell while adapting to aging effects, thereby achieving strong predictive performance.

    \item The proposed model is validated on three public battery datasets: (i) a dataset with a constant discharge current across all cells and charge-discharge cycles, (ii) a dataset with fixed discharge profiles across cycles within each cell but differing between cells, and (iii) a dataset with random discharge profiles across both cycles and cells. Unlike \cite{wang2023inherently}, where training and testing data are taken from the same cell, our data-splitting strategy uses separate cells for training, validation, and testing. This approach provides a more general and fair evaluation of model performance. Under this setting, SeqBattNet outperforms strong baselines. In this work, we do not consider temperature effects.
\end{itemize}

The rest of this paper is structured as follows. Section~\ref{sec:preliminaries} introduces the preliminaries. 
Section~\ref{sec:proposed_method} presents the proposed approach, including model architecture, and evaluation metrics. Section~\ref{sec:dataset} details the datasets and provides data analysis. Section~\ref{sec:experiments} describes the experimental setup, reports model performance, and discusses future directions. Finally, Section~\ref{sec:conclusion} summarizes the findings.

\section{Preliminaries}\label{sec:preliminaries}
The Gated Recurrent Unit (GRU)~\cite{dey2017gate} is widely used in sequence modeling~\cite{nosouhian2021review}, as it effectively captures temporal dependencies while reducing parameter complexity compared to Long Short-Term Memory (LSTM) networks~\cite{hochreiter1997long} and Transformers~\cite{vaswani2017attention}. Given input $\mathbf{x}_t \in \mathbb{R}^d$ and previous hidden state 
$\mathbf{h}_{t-1} \in \mathbb{R}^m$, a GRU cell updates as:
\begin{align}
\mathbf{z}_t &= \sigma\!\left(\mathbf{W}_z \mathbf{x}_t + \mathbf{U}_z \mathbf{h}_{t-1} + \mathbf{b}_z \right), \tag{update gate}\\
\mathbf{r}_t &= \sigma\!\left(\mathbf{W}_r \mathbf{x}_t + \mathbf{U}_r \mathbf{h}_{t-1} + \mathbf{b}_r \right), \tag{reset gate}\\
\tilde{\mathbf{h}}_t &= \tanh\!\left(\mathbf{W}_h \mathbf{x}_t + \mathbf{U}_h \left(\mathbf{r}_t \odot \mathbf{h}_{t-1}\right) + \mathbf{b}_h \right), \tag{candidate state}\\
\mathbf{h}_t &= (1-\mathbf{z}_t)\odot \mathbf{h}_{t-1} + \mathbf{z}_t \odot \tilde{\mathbf{h}}_t, \tag{new hidden state}
\end{align}
where $\sigma(\cdot)$ is the sigmoid function, $\tanh(\cdot)$ is the hyperbolic tangent, and $\odot$ denotes element-wise multiplication. The variables are defined as follows: 
$\mathbf{x}_t \in \mathbb{R}^d$ is the input vector at time step $t$, 
$\mathbf{h}_{t-1}, \mathbf{h}_t \in \mathbb{R}^m$ are the previous and current hidden states, 
$\mathbf{z}_t, \mathbf{r}_t, \tilde{\mathbf{h}}_t \in \mathbb{R}^m$ are the update gate, reset gate, and candidate state, respectively, 
$\mathbf{W}_{(\cdot)} \in \mathbb{R}^{m \times d}$ and $\mathbf{U}_{(\cdot)} \in \mathbb{R}^{m \times m}$ are learnable weight matrices, 
and $\mathbf{b}_{(\cdot)} \in \mathbb{R}^m$ are bias terms. The GRU combines the update gate $\mathbf{z}_t$ and reset gate $\mathbf{r}_t$ to control information flow and capture temporal dependencies while avoiding vanishing gradients. In this work, we use the GRU cell, denoted as $\mathrm{GRU}(\cdot,\cdot)$, in the encoder component of our proposed model.

\section{Proposed Method}\label{sec:proposed_method}
\begin{figure}[H]
  \centering
  \includegraphics[width=\textwidth]{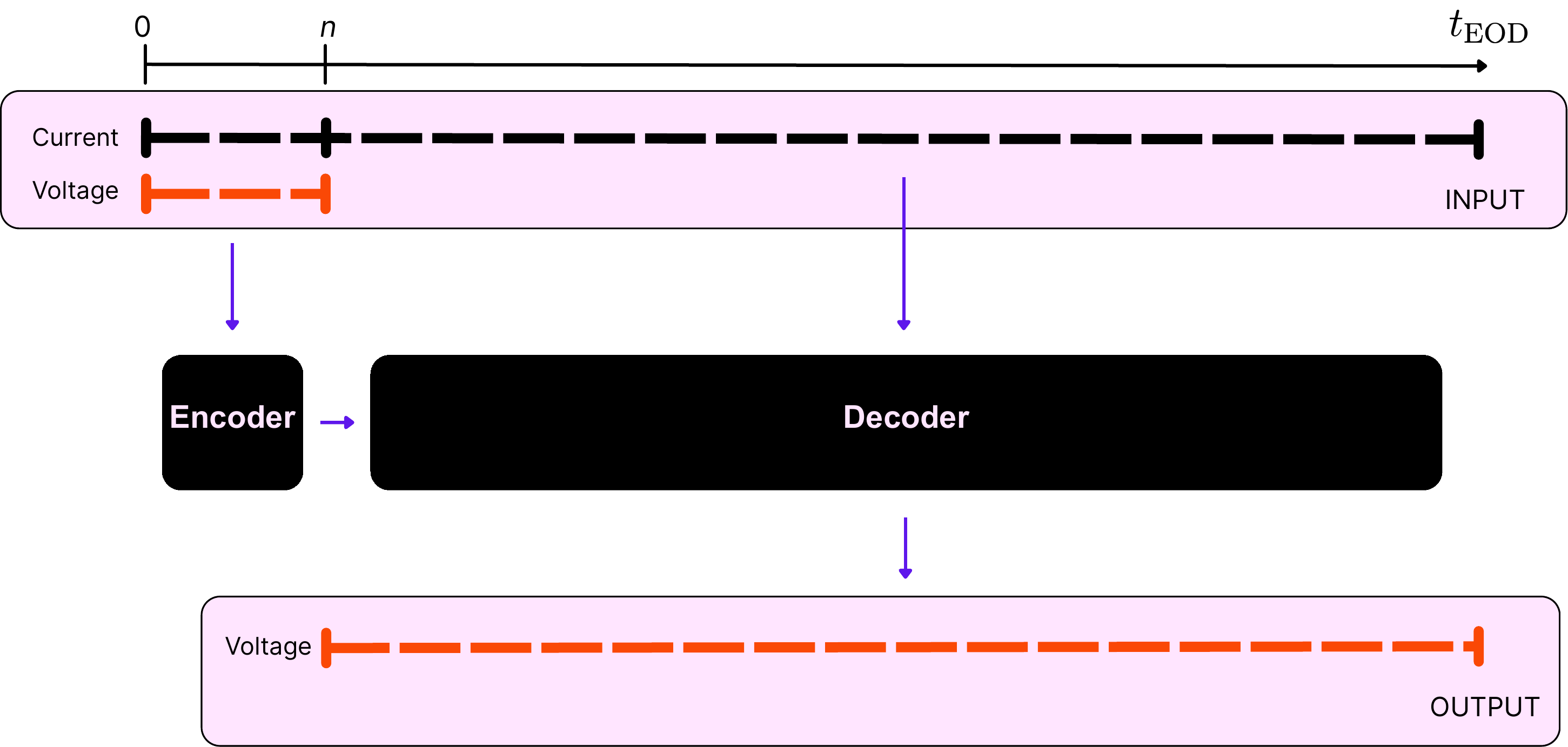}
  \caption{Overall architecture of our proposed SeqBattNet.}
  \label{fig:Overall_flow}
\end{figure}

Our SeqBattNet consists of two main components: an encoder and a decoder, as illustrated in Figure~\ref{fig:Overall_flow}. 
The encoder employs a novel HRM--GRU to predict adaptation parameters corresponding to the current battery state. The decoder implements a physics-informed ECM that combines physical equations with neural networks to predict the terminal voltage from the estimated adaptation parameters and the input current. SeqBattNet is trained with a customized loss function that places stronger emphasis on the beginning and the end of the predicted voltage trajectory, thereby guiding the model to better follow the true terminal voltage sequence.


\subsection{Encoder: Adaptation Parameter Prediction}\label{subsec:encoder}

Our encoder leverages the Hierarchical Reasoning Model (HRM)~\cite{wang2025hierarchical} to design a HRM--GRU encoder. HRM provides a multi-level reasoning framework that enables effective representation of complex temporal dependencies in the battery state during each cycle. Whereas \cite{wang2025hierarchical} employs Transformers for hierarchical reasoning, we replace Transformers with GRUs. This substitution reduces computational complexity and improves prediction accuracy, as demonstrated in the experiment section, making the model more suitable for real-world applications. For each discharge cycle, the HRM--GRU encoder is provided with an initial segment of current and voltage, from the start of discharge up to a fixed length of $n$ time steps, to estimate the battery adaptation parameters: the ohmic resistance $R_0$ (capturing the instantaneous voltage drop), the RC branch time constants $\boldsymbol{\tau} = \{\tau_1,\dots,\tau_{e_{\mathrm{rc}}}\}$ (describing diffusion and polarization dynamics), the initial state of charge $\mathrm{SOC}_0$, the state of health $\mathrm{SOH}$ (indicating capacity degradation at the current discharge cycle), and the initial RC branch voltages $\mathbf{v}_{RC,0}$. In this work, we set the number of RC branches to $e_{\mathrm{rc}}=2$. Detailed architecture of HRM--GRU encoder block is represented as follows.

At each discharge cycle, the initial sequences of current $I_{init} \in \mathbb{R}^n$ and voltage $V_{init} \in \mathbb{R}^n$ are combined into $\mathbf{x} = [I_{init}, V_{init}] \in \mathbb{R}^{n \times 2}$, which serves as the input to the HRM--GRU encoder. Here, $n$ denotes the sequence length. The HRM--GRU first maps the input $\mathbf{x}$ into an embedding space through a linear layer:  
\[
\mathbf{e} = \mathbf{W}_{\mathrm{emb}}\mathbf{x} + \mathbf{b}_{\mathrm{emb}} \;\in\; \mathbb{R}^{n \times d_{\mathrm{emb}}},
\]
where $d_{\mathrm{emb}}$ denotes the embedding dimension. 
$\mathbf{W}_{\mathrm{emb}} \in \mathbb{R}^{2 \times d_{\mathrm{emb}}}$ is the weight matrix of the linear layer, 
and $\mathbf{b}_{\mathrm{emb}} \in \mathbb{R}^{d_{\mathrm{emb}}}$ is the bias vector.

The core component of the hierarchical reasoning architecture, comprising the preparatory micro-update and main update phases, is used to process the embedding $\mathbf{e}$. We begin with the preparatory micro-update phase. Let $\mathbf{z}^L_s \in \mathbb{R}^{d_L}$ and $\mathbf{z}^H_s \in \mathbb{R}^{d_H}$ denote the low- and high-level hidden states at sequence index $s = 0,\dots,n$ along the time dimension of the input $\mathbf{x}$, with initial states $\mathbf{z}^L_0 = \mathbf{0}$ and $\mathbf{z}^H_0 = \mathbf{0}$. Here, $d_L$ and $d_H$ represent the dimensionalities of the low- and high-level latent spaces, respectively. Before consuming $\mathbf{e}_{s}$, we perform $Q = NT - 1$ gradient-free micro-updates, where $N$ denotes the number of reasoning layers and $T$ denotes the number of update steps per layer, in order to obtain the pre-states 
$(\mathbf{z}^L_{s^-}, \mathbf{z}^H_{s^-})$. We formalize this process through the gradient-free micro-update operator 
$\mathcal{U}_{N,T}(\cdot)$, defined as
\[
(\mathbf{z}^L_{s^-}, \mathbf{z}^H_{s^-})
= \mathcal{U}_{N,T}\!\big(\mathbf{z}^L_{s-1}, \mathbf{z}^H_{s-1};\, \mathbf{e}_{s}\big),
\]
where $Q = N T - 1$ denotes the total number of micro-updates. 
Specifically, for $q = 1, \dots, Q$,
\begin{align*}
\mathbf{z}^L_{(q)} &= \mathrm{GRU}_L\!\big([\mathbf{e}_{s},\, \mathbf{z}^H_{(q-1)}],\, \mathbf{z}^L_{(q-1)}\big), \\[6pt]
\mathbf{z}^H_{(q)} &= 
\begin{cases}
\mathrm{GRU}_H\!\big(\mathbf{z}^L_{(q)},\, \mathbf{z}^H_{(q-1)}\big), & \text{if } q \equiv 0 \pmod T, \\[6pt]
\mathbf{z}^H_{(q-1)}, & \text{otherwise},
\end{cases}
\end{align*}
with assignment 
\[
(\mathbf{z}^L_{s^-}, \mathbf{z}^H_{s^-}) = (\mathbf{z}^L_{(Q)}, \mathbf{z}^H_{(Q)}).
\]
Where $\mathbf{z}^L_{s^-}$ and $\mathbf{z}^H_{s^-}$ denote the low- and high-level hidden states obtained after $Q$ preparatory micro-updates.

In the subsequent main update step, the current input embedding $\mathbf{e}_{s}$ is integrated into the hierarchical states:
\begin{align*}
\mathbf{z}^L_s &= \mathrm{GRU}_L\!\big([\mathbf{e}_{s},\,\mathbf{z}^H_{s^-}],\,\mathbf{z}^L_{s^-}\big),\\
\mathbf{z}^H_s &= \mathrm{GRU}_H\!\big(\mathbf{z}^L_s,\,\mathbf{z}^H_{s^-}\big).
\end{align*}
Here, $\mathbf{z}^L_s$ and $\mathbf{z}^H_s$ represent the updated hidden states after incorporating $\mathbf{e}_{s}$ through the low-to-high GRU updates. The notation $[\cdot,\cdot]$ indicates feature concatenation. A dropout layer is then applied to $\mathbf{z}^H_s$ for prediction. After completing the high-level update at step $s$, we apply layer normalization to the hidden state $\mathbf{z}^H_s$, followed by a linear projection: 
\[
\mathbf{o}_s \;=\; \mathbf{W}_{\text{out}}\,\mathrm{Norm}(\mathbf{z}^H_s) + \mathbf{b}_{\text{out}}
\;\in\; \mathbb{R}^{d_{\text{out}}},
\]
where $d_{\text{out}} = 2\,e_{\mathrm{rc}} + 3$, and $\mathrm{Norm}(\cdot)$ denotes layer normalization~\cite{ba2016layer}. Stacking across the $n$ time steps yields $[\mathbf{o}_1,\dots,\mathbf{o}_n] \in \mathbb{R}^{n \times d_{\text{out}}}$.
We use the final vector as output, yielding:
\[
\big[R_0,\;\boldsymbol{\tau},\; \mathrm{SOC}_0,\; \mathrm{SOH},\; \mathbf{w}\big] = \mathbf{o}_n 
\]
where
$R_0 \in \mathbb{R}$, 
$\boldsymbol{\tau} \in \mathbb{R}^{\,e_{\mathrm{rc}}}$, 
$\mathrm{SOC}_0 \in \mathbb{R}$, 
$\mathrm{SOH} \in \mathbb{R}$, and 
$\mathbf{w} \in \mathbb{R}^{\,e_{\mathrm{rc}}}$. 
To ensure that these parameters remain physically valid, range-safe mappings are applied as follows:
\begin{align*}
R_0 &\;\mapsto\; \mathrm{Aff}\text{-}\sigma(R_0;\,10^{-3},\,0.5), \\[4pt]
\boldsymbol{\tau} &\;\mapsto\; \mathrm{Aff}\text{-}\sigma(\boldsymbol{\tau};\,10^{-2},\,10^{5}), \\[4pt]
\mathrm{SOC}_0 &\;\mapsto\; \sigma(\mathrm{SOC}_0), \\[4pt]
\mathrm{SOH} &\;\mapsto\; \sigma(\mathrm{SOH}), \\[4pt]
\mathbf{w} &\;\mapsto\; \mathrm{Softmax}(\mathbf{w}).
\end{align*}
where $\mathrm{Aff}\text{-}\sigma(z;\ell,h) = \ell + (h-\ell)\,\sigma(z)$. $\mathrm{Aff}\text{-}\sigma$ is an affine-sigmoid mapping that constrains a parameter to a prespecified range $[\ell,h]$, 
$\sigma$ is the standard logistic sigmoid mapping to $(0,1)$, and $\mathrm{Softmax}$ ensures normalized non-negative weights.  
The specific bounds ($10^{-3}, 0.5$ for $R_0$ and $10^{-2}, 10^5$ for $\boldsymbol{\tau}$) are chosen based on typical ranges of ohmic resistance~\cite{somakettarin2019characterization} and RC time constants observed in lithium-ion cells~\cite{sakile2022estimation}, thereby enforcing physical plausibility. The vector $\mathbf{v}_{RC,0} \in \mathbb{R}^{\,e_{\mathrm{rc}}}$, which provides the initial RC branch voltages, is computed as 
\begin{equation}
\begin{aligned}
\mathrm{OCV}_0 &\;=\; V_{\mathrm{EOD}} + (V_0 - V_{\mathrm{EOD}})\,\sigma(g(\mathrm{SOC}_0))\\
s_0 &\;=\; \mathrm{OCV}_0\;-\; R_0 I_{\text{last}}\;-\; V_{\text{last}}, \\
\mathbf{v}_{RC,0} &\;=\; \mathbf{w}\, s_0 \in \mathbb{R}^{\,e_{\mathrm{rc}}},
\end{aligned}
\label{eq:init_rc}
\end{equation}
where $I_{\text{last}}$ and $V_{\text{last}}$ denote the last observed current and voltage in the input window; $V_0$ and $V_{\mathrm{EOD}}$ represent the nominal open-circuit voltage and the end-of-discharge cutoff voltage, respectively; $\mathrm{OCV}_0$ is the open-circuit voltage at time step $n+1$; $\mathbf{w}$ denotes the weights associated with each RC branch; and $g(\cdot)$ is a feedforward neural network. 
Equation~\ref{eq:init_rc} will be explained in the next subsection~\ref{subsec:decoder}.

In each discharge cycle, the encoder first predicts the initial parameters 
$R_0$, $\boldsymbol{\tau}$, $\mathrm{SOC}_0$, $\mathrm{SOH}$, and $\mathbf{v}_{RC,0}$, 
which are then passed to the decoder to predict the terminal voltage for each input current. The hyperparameters of HRM--GRU are summarized in Table~\ref{tab:HRM_GRU}. They are obtained using grid search~\cite{liashchynskyi2019grid}. The parameters $N$ and  $T$ are searched within the range \{1, 2, 3\}. We restrict $N$ and $T$ to small values in order to maintain the real-time prediction requirement in real-world applications. $d_{emb}$, $d_{L}$ and $d_{H}$ are searched over the set \{16, 32, 64, 128, 256\}. The operation of the decoder is presented in the next subsection.

\begin{table}[H]
\centering
\caption{HRM--GRU Hyperparameters}
\begin{tabular}{lc}
\toprule
\textbf{Parameter} & \textbf{Value} \\
\midrule
$N$            & 1 \\
$T$            & 2 \\
$d_{\mathrm{emb}}$ & 32 \\
$d_L$ & 128 \\
$d_H$ & 64 \\
\bottomrule
\end{tabular}
\label{tab:HRM_GRU}
\end{table}

\subsection{Decoder: Terminal Voltage Prediction}
\label{subsec:decoder}

The decoder implements a physics-informed ECM, inspired by the traditional first-order ECM~\cite{tran2021comparative}, in which feedforward neural networks (FNNs) are employed to predict the open-circuit voltage and the resistances of the RC branches. The adaptation parameters predicted by the encoder block, are then used to estimate the terminal voltage at each time step based on the input current. 

\begin{algorithm}[H]
\caption{Decoder: Physics-Informed ECM Update for $m = n+1, \dots, t_{\mathrm{EOD}}$}
\label{alg:decoder_ecm}
\begin{algorithmic}[1]
\REQUIRE Adaptation parameters: $R_0$, $\boldsymbol{\tau}$, $\mathrm{SOC}_0$, $\mathrm{SOH}$, $\mathbf{v}_{RC,0}$, and  the current input sequence $\mathbf{I}$.
\STATE Compute decay coefficient: $\alpha = \exp(-\Delta t / \boldsymbol{\tau})$
\FOR{$m = n+1$ to $t_{\mathrm{EOD}}$}
    \STATE \textbf{OCV prediction:} 
        \begin{equation}
        \mathrm{OCV}_m = V_{\mathrm{EOD}} + (V_0 - V_{\mathrm{EOD}})\,\sigma(g(\mathrm{SOC}_m))\label{eq:1}
        \end{equation}
    \STATE \textbf{RC branch update:}
        \begin{equation}
        \begin{aligned}
        \mathbf{r}_{RC,m} &= r_{\min} + (r_{\max}-r_{\min})\,\sigma\!\big(f(\mathrm{SOC}_m, \mathrm{SOH})\big), \\
        \mathbf{v}_{RC,m} &= \alpha \mathbf{v}_{RC,m-1} + (1-\alpha)\,\mathbf{r}_{RC,m} I_m.
        \end{aligned}\label{eq:2}
        \end{equation}
        
    \STATE \textbf{SOC update:}
        \begin{equation}
        \mathrm{SOC}_{m+1} = \Pi_{[0,1]}\!\left( \mathrm{SOC}_m - \frac{I_m \Delta t}{3600\,C_{\mathrm{eff}}} \right)\label{eq:3}
        \end{equation}
    \STATE \textbf{Terminal voltage prediction:}
        \begin{equation}
        \widehat{V}_m = \mathrm{OCV}_m - R_0 I_m - \sum_{k=1}^{e_{\mathrm{rc}}} v_{RC,m,k}\label{eq:4}
        \end{equation}
\ENDFOR
\ENSURE Predicted voltage sequence:
\begin{equation}
\widehat{\mathbf{V}} = \big[ V_0,\dots,V_{n},\; \widehat{V}_{n+1}, \dots, \widehat{V}_{t_{\mathrm{EOD}}} \big]\label{eq:5}
\end{equation}
\end{algorithmic}
\end{algorithm}
The overall algorithm of the decoder block is presented in Algorithm~\ref{alg:decoder_ecm}. 
The initial parameters from the encoder block, 
$R_0 \in \mathbb{R}$, 
$\boldsymbol{\tau} \in \mathbb{R}^{\,e_{\mathrm{rc}}}$, 
$\mathrm{SOC}_0 \in \mathbb{R}$, 
$\mathrm{SOH} \in \mathbb{R}$, 
and $\mathbf{v}_{RC,0} \in \mathbb{R}^{\,e_{\mathrm{rc}}}$, 
together with the decay coefficient $\alpha$ and the current sequence 
$\mathbf{I} = \{ I_{n+1}, I_{n+2}, \dots, I_{t_{\mathrm{EOD}}} \}$, 
where each $I_m \in \mathbb{R}$ is positive for discharge and negative for charge, 
are used as inputs to the decoder to generate the terminal voltage at each time step 
$m = n+1, n+2, \dots, t_{\mathrm{EOD}}$. 

Here, $t_{\mathrm{EOD}}$ denotes the step at which the battery reaches the EOD threshold. 
The decay coefficient $\alpha$ is defined as 
$\alpha = \exp(-\Delta t / \boldsymbol{\tau})$, 
which serves as a discrete-time decay factor that captures how the RC branch voltage 
relaxes over each sampling interval $\Delta t$. 
Both $\boldsymbol{\tau}$ and $\Delta t$ are expressed in seconds.

In each time step $m$, the open-circuit voltage $OCV_m \in \mathbb{R}$ is modeled by a feedforward neural network $g(\cdot)$ that maps the $SOC_m$ into the voltage range $[V_{\mathrm{EOD}}, V_0]$, as defined in Eq.~\ref{eq:1}. $V_0$ denotes the nominal open-circuit voltage at full charge, and
$V_{\mathrm{EOD}}$ denotes the end-of-discharge cutoff voltage. The OCV represents the equilibrium voltage of the battery at no load, linking SOC to the intrinsic electrochemical potential. 

For the RC branches, the resistance $\mathbf{r}_{RC,m} \in \mathbb{R}^{\,e_{\mathrm{rc}}}$ is modeled by a feedforward neural network $f(\cdot,\cdot)$ that takes $\mathrm{SOC}_m$ and $\mathrm{SOH}$ as inputs and maps them into the range $[r_{\mathrm{min}},\, r_{\mathrm{max}}]$, as defined in Eq.~\ref{eq:2}. In our study, we set $r_{\mathrm{min}} = 10^{-4}$ and $r_{\mathrm{max}} = 1$ as physically recommended constraints following \cite{somakettarin2019characterization}. The $\mathbf{r}_{RC,m}$ is used to compute the branch voltage $\mathbf{v}_{RC,m}$ at each time step. The $\mathbf{v}_{RC,m} \in \mathbb{R}^{\,e_{\mathrm{rc}}}$ are calculated based on the decay coefficient $\alpha$, the estimated $\mathbf{r}_{RC,m}$, and the current $I_m$. These voltages represent the dynamic relaxation behavior of the RC branches, capturing the delayed response of the battery to current excitation.

For the SOC update, $SOC_{m+1}\in \mathbb{R}$ is computed to prepare for the next time step $m+1$, as defined in Eq.~\ref{eq:3}. Note that $\mathrm{SOC}_0$, predicted by the encoder, refers to the SOC at time step $m=n+1$. The operator $\Pi_{[0,1]}$ denotes clipping to the interval $[0,1]$. The effective capacity $C_{\mathrm{eff}}$ is calculated by SOH (predicted by the encoder) and defined as
\[
C_{\mathrm{eff}} = \beta C_{\mathrm{EOL}} 
+ \big(C_{\mathrm{rated}} - \beta C_{\mathrm{EOL}}\big)\,\mathrm{SOH},
\]
where $C_{\mathrm{rated}}$ denotes the nominal rated capacity, and $C_{\mathrm{EOL}}$ is the end-of-life capacity, typically set to $0.8\,C_{\mathrm{rated}}$. 
Since terminal voltage prediction begins after the initial segment processed by the encoder, the encoder input the initial current segment that consumes a portion of the battery capacity. 
To account for capacity fade, we set $\beta = 0.8$ as the fractional capacity threshold. The detailed architectures of $g(\cdot)$ and $f(\cdot)$ are shown in Table~\ref{tab:fnn_arch}. The hyperparameters in Table~\ref{tab:fnn_arch} are obtained using grid search over the values $\{16, 32, 64, 128, 256\}$.

\begin{table}[H]
\centering
\caption{FNN hyperparameters.}
\begin{tabular}{l l c c}
\toprule
\textbf{Network} & \textbf{Layer} & \textbf{Input} & \textbf{Output} \\
\midrule
$g(\cdot)$ & Linear+SiLU    & 1  & 32 \\
           & Linear+SiLU    & 32 & 32 \\
           & Linear+Sigmoid & 32 & 1 \\
\midrule
$f(\cdot)$ & Linear+SiLU    & 2 & 32 \\
           & Linear+Sigmoid & 32 & $e_{\mathrm{rc}}$ \\
\bottomrule
\end{tabular}
\label{tab:fnn_arch}
\end{table}

For terminal voltage prediction, $\widehat{V}_m \in \mathbb{R}$ is computed from the estimated $OCV_m$, 
the estimated $\mathbf{v}_{RC,m}$, the estimated $R_0$, and the current $I_m$, as defined in Eq.~\ref{eq:4}. This ensures that the predicted terminal voltage accounts for both the instantaneous resistive drop across $R_0$ and the dynamic relaxation behavior of the RC branches.

Finally, the predicted voltage sequence $\widehat{\mathbf{V}}$ concatenates the measured past with the predicted future, as defined in Eq.~\ref{eq:5}. The predicted terminal voltage is optimized using the proposed loss function, which is introduced in the next subsection.

\subsection{Loss Function} 
We propose a weighted loss function $\mathcal{L}_{\mathrm{wL1}}$ based on the Huber loss~\cite{girshick2015fast} $\ell_{\beta}$ with adaptive weighting across the sequence. 
This formulation combines the robustness of the $\ell_1$ loss against outliers with the stability of the $\ell_2$ loss near zero error, 
while additionally emphasizing more critical time steps within the sequence. 
For a sequence of length $t_{\mathrm{EOD}}$, the loss is defined as
\[
\mathcal{L}_{\mathrm{wL1}}
= \frac{1}{\sum_{i=1}^{t_{\mathrm{EOD}}} a_i w_i}
\sum_{i=1}^{t_{\mathrm{EOD}}} a_i \, w_i \, \ell_{\beta}\!\left( \hat{V}_i, V_i \right),
\]
where \(V_i\) and \(\hat{V}_i\) denote the true and predicted terminal voltages at time step \(i\),
\(a_i \in \{0,1\}\) is a masking indicator, and \(w_i\) is the adaptive weight. In the training process, all sequences are zero-padded to a common length, and a mask $a$ is applied to restrict computations to the original valid entries. This procedure allows the model to be trained on batches of samples simultaneously rather than processing individual sequences sequentially. The Huber loss $\ell_\beta$ is given by
\[
\ell_{\beta}(\hat{V}_i,V_i) =
\begin{cases}
\frac{1}{2\beta}(V_i - \hat{V}_i)^2, & |V_i - \hat{V}_i| < \beta, \\[6pt]
|V_i - \hat{V}_i| - \tfrac{1}{2}\beta, & \text{otherwise},
\end{cases}
\]
with $\beta = 0.1$ in our implementation. The weight sequence $\{w_1, w_2, \dots, w_{t_{\mathrm{EOD}}}\}$ is designed to emphasize earlier prediction steps and the last prediction step. Specifically, for a subsequence of length $t_{\mathrm{EOD}}$, the weights are constructed as
\[
w_i = \tfrac{1}{2}\Big( \, \underbrace{\Big(10 - \tfrac{9}{t_{\mathrm{EOD}}-1}(i-1)\Big)}_{\text{linear schedule from 10 to 1}} 
+ \delta_{i,t_{\mathrm{EOD}}}\cdot 30 \Big), 
\qquad i=1,\dots,t_{\mathrm{EOD}},
\]
where the first term decreases linearly from $10$ at $i=1$ to $1$ at $i=t_{\mathrm{EOD}}$, and $\delta_{i,t_{\mathrm{EOD}}}$ is an indicator that adds a large weight ($30$) to the last prediction step. 
The values $10$ and $30$ were chosen empirically after testing several candidates $\{5,10,20,30,40\}$. By emphasizing both the beginning and the end of the sequence, this weighting scheme guides the model to follow the overall discharge profile more accurately.

\subsection{Evaluation Metrics}
To evaluate the performance of battery voltage prediction, we employ three commonly used metrics: Root Mean Square Error (\(RMSE\)), Mean Absolute Error (\(MAE\)), and Mean Absolute Percentage Error (\(MAPE\)), as referenced in \cite{wang2023inherently}. 
These metrics are defined as follows:

\[
RMSE(\mathbf{V}, \widehat{\mathbf{V}}) = 
\sqrt{\frac{1}{t_{EOD}} \sum_{i=1}^{t_{EOD}} \big(V_i - \widehat{V}_i\big)^2},
\]

\[
MAE(\mathbf{V}, \widehat{\mathbf{V}}) = 
\frac{1}{t_{EOD}} \sum_{i=1}^{t_{EOD}} \big|V_i - \widehat{V}_i\big|,
\]

\[
MAPE(\mathbf{V}, \widehat{\mathbf{V}}) = 
\frac{100}{t_{EOD}} \sum_{i=1}^{t_{EOD}} \frac{\big|V_i - \widehat{V}_i\big|}{V_i}.
\]
Here, \(V_i\) denotes the measured terminal voltage at time step \(i\), and \(\widehat{V}_i\) denotes the corresponding predicted voltage. Better predictive performance is indicated by lower values of \(RMSE\), \(MAE\), and \(MAPE\).

\section{Dataset}\label{sec:dataset}
\paragraph{TRI Dataset~\cite{severson2019data} - Same discharge load profiles across cells and cycles}
The dataset includes 124 LFP/graphite A123 APR18650M1A LIBs, each with a nominal capacity of 1.1\,Ah and a nominal open-circuit voltage of 3.6\,V. While various charge protocols are applied, all cells share identical discharge load profiles across cycles, ensuring highly consistent operating conditions. Cycle lives span from 150 to 2,300 cycles. Cells were discharged at a constant current of 4C down to 2\,V, followed by a constant-voltage phase at 2\,V until the current declined to C/50. Charging was conducted at rates between 3.6C and 6C under a controlled temperature of 30\,\textdegree C. With approximately 96,700 recorded cycles, this is among the largest publicly available datasets for studying fast-charging behavior. In this dataset, cell \texttt{b1c0} is used for training, \texttt{b2c13} for validation, and cells \texttt{b1c4} and \texttt{b1c2} are reserved for testing. Accordingly, we set the model parameters to \(C_{\mathrm{rated}} = 1.1\,\mathrm{Ah}\), \(C_{\mathrm{EOL}} = 0.88\,\mathrm{Ah}\) (i.e., \(0.8\,C_{\mathrm{rated}}\)), \(V_{0} = 3.6\,\mathrm{V}\), and \(V_{\mathrm{EOD}} = 2.0\,\mathrm{V}\).

\paragraph{RT-Batt Dataset~\cite{ma2022real} - Different profiles across cells, same across cycles} 
The dataset consists of 77 LFP/graphite A123 APR18650M1A cells (1.1 Ah, 3.6 V nominal OCV), each subjected to a unique multi-stage discharge profile that remains consistent across cycles. All cells followed an identical fast-charging protocol and were tested in thermostatic chambers at 30\,\textdegree C. The dataset includes 146,122 discharge cycles, making it one of the largest for studying varied discharge behaviors. Cycle life ranges from 1,100 to 2,700 cycles. In this dataset, cell \texttt{1-1} is used for training, \texttt{1-4} for validation, and cells \texttt{1-2} and \texttt{1-3} are used for testing. Accordingly, we set the model parameters to \(C_{\mathrm{rated}} = 1.1\,\mathrm{Ah}\), \(C_{\mathrm{EOL}} = 0.88\,\mathrm{Ah}\) (i.e., \(0.8\,C_{\mathrm{rated}}\)), \(V_{0} = 3.6\,\mathrm{V}\), and \(V_{\mathrm{EOD}} = 2.0\,\mathrm{V}\).

\paragraph{NASA Dataset~\cite{bole2014adaptation} - Different profiles across both cells and cycles} 
This dataset features high variability in discharge profiles across different cells and cycles, closely simulating real-world battery usage. Provided by the NASA Ames Prognostics Data Repository, it includes four 18650 lithium-ion cells (RW3–RW6), each charged at a constant 2\,A to 4.2\,V and discharged to 3.2\,V using randomized current profiles between 0.5\,A and 4\,A. This variability makes the dataset ideal for testing model robustness under diverse operating conditions. In this dataset, cell \texttt{RW3} is used for training, \texttt{RW6} for validation, and cells \texttt{RW4} are \texttt{RW5} are used for testing. Accordingly, we set the model parameters to \(C_{\mathrm{rated}} = 2.22\,\mathrm{Ah}\), \(C_{\mathrm{EOL}} = 1.33\,\mathrm{Ah}\) (i.e., \(0.6\,C_{\mathrm{rated}}\)), \(V_{0} = 4.2\,\mathrm{V}\), and \(V_{\mathrm{EOD}} = 3.2\,\mathrm{V}\).

\begin{figure}[H]
    \centering
    \begin{subfigure}[b]{0.48\textwidth}
        \centering
        \includegraphics[width=\textwidth]{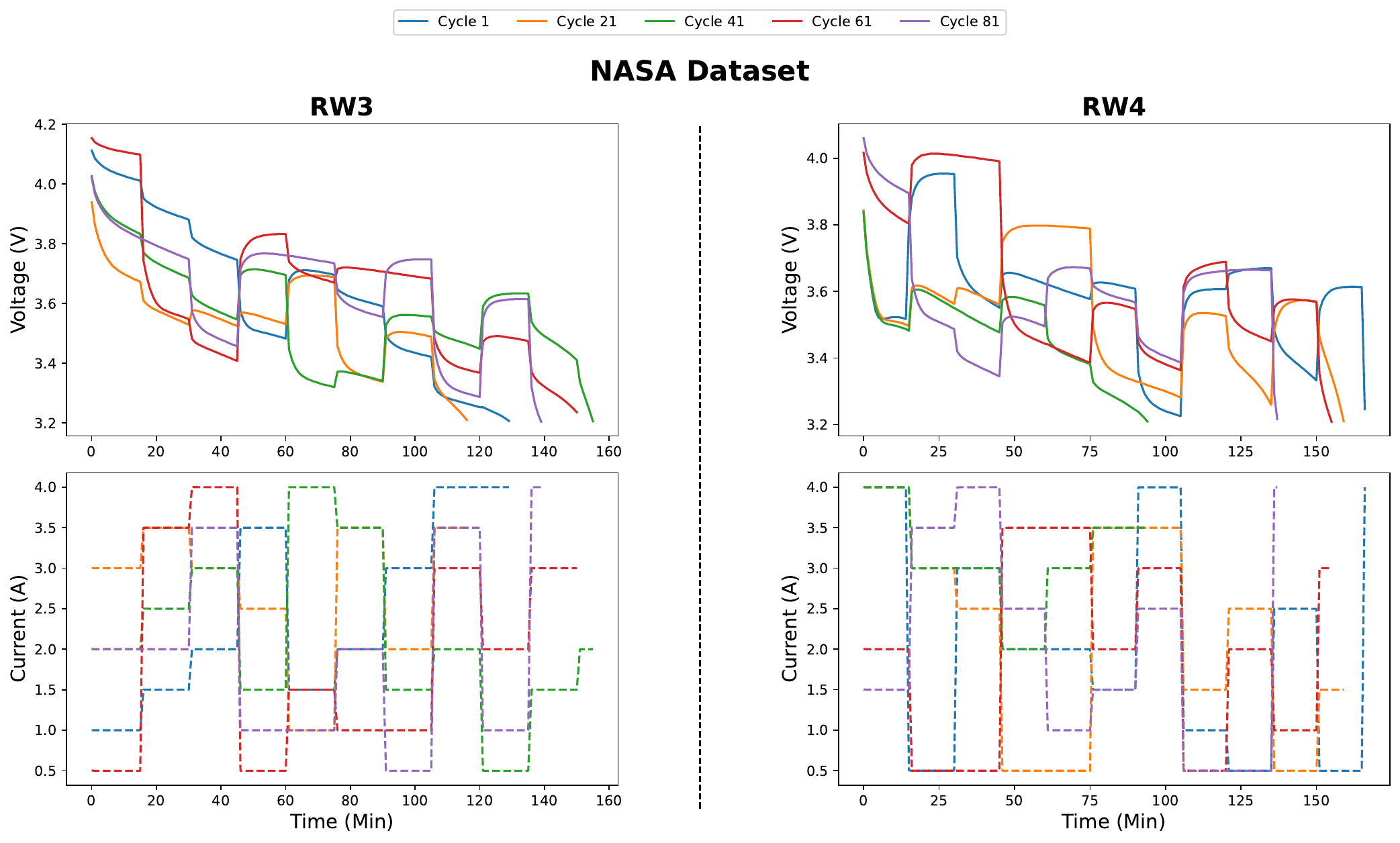}
        \label{fig:image1}
    \end{subfigure}
    \begin{subfigure}[b]{0.48\textwidth}
        \centering
        \includegraphics[width=\textwidth]{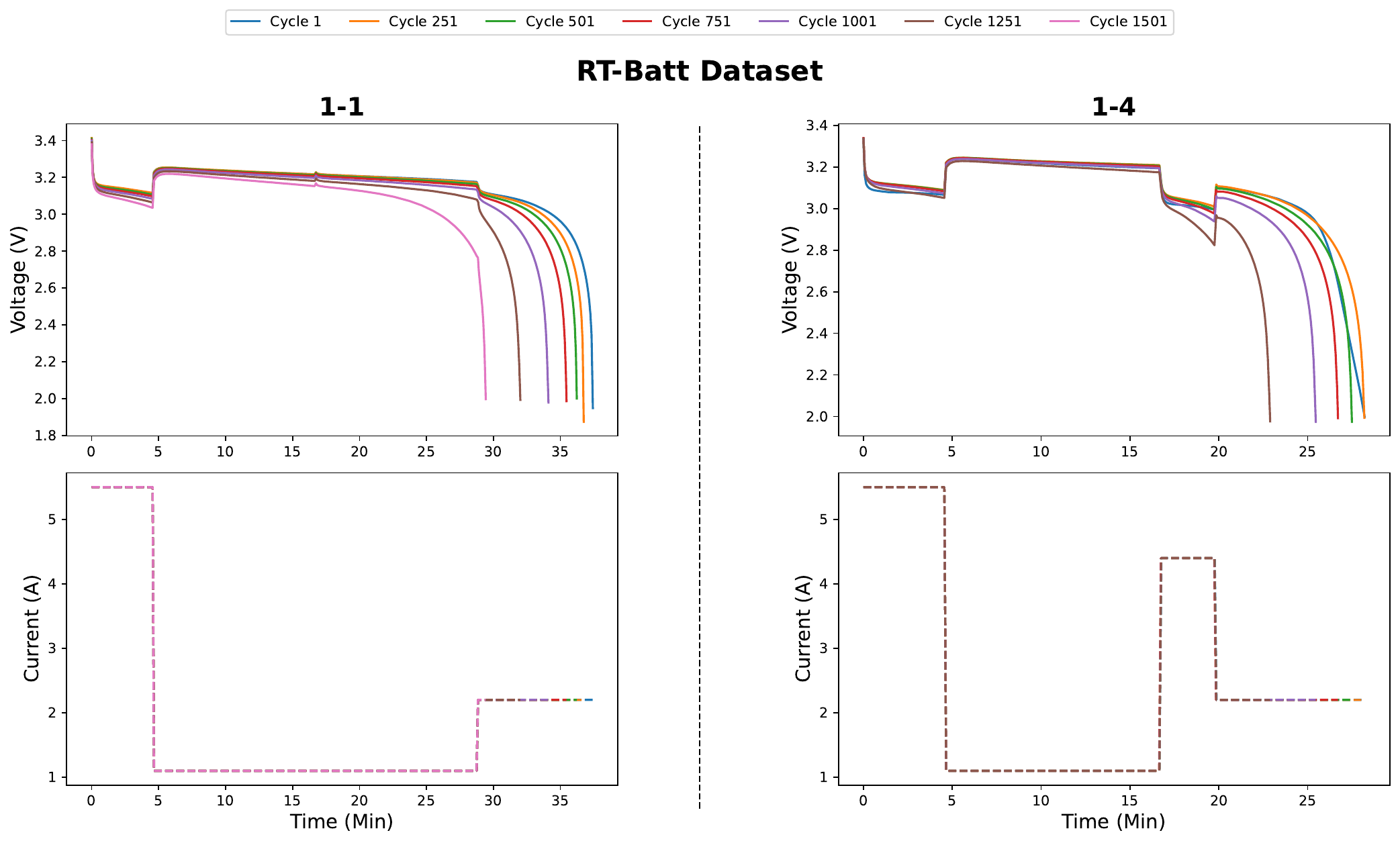}
        \label{fig:image2}
    \end{subfigure}

    \begin{subfigure}[b]{0.5\textwidth}
        \centering
        \includegraphics[width=\textwidth]{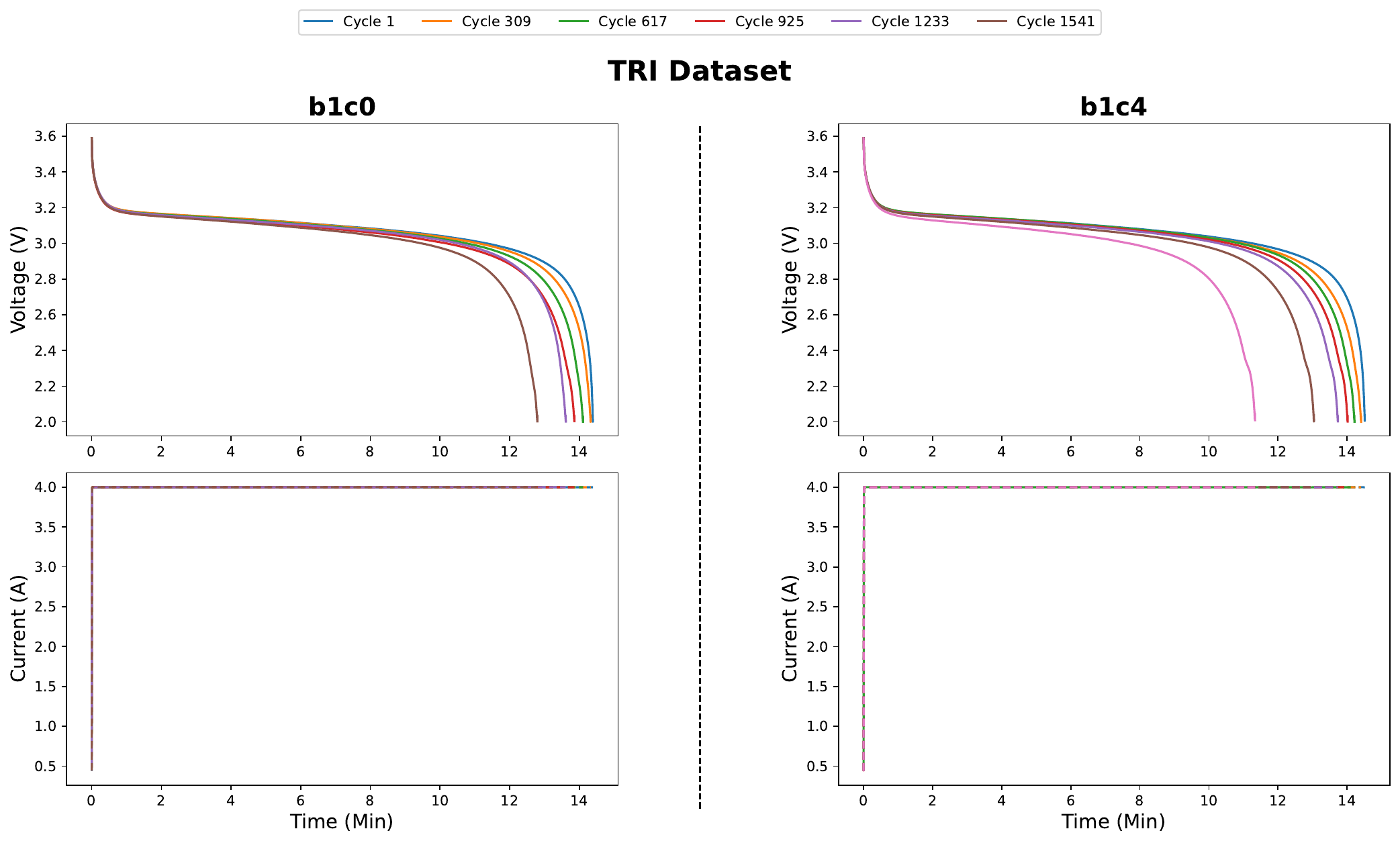}
        \label{fig:image3}
    \end{subfigure}

    \caption{Visualization of the battery datasets: NASA, RT-Batt, and TRI}
    \label{fig:3_dataset_overall}
\end{figure}

Figure~\ref{fig:3_dataset_overall} provides an overall visualization of the discharge process across several cycles, from beginning of life to end of life, for three datasets. As batteries age, the voltage tends to decline more rapidly toward the end of discharge, that is particularly evident in the TRI and RT-Batt datasets. Each dataset exhibits different characteristics: the TRI dataset has identical discharge load profiles across cells and cycles; 
the RT-Batt dataset has different profiles across cells but identical profiles across cycles; 
and the NASA dataset has different profiles across both cells and cycles. These variations simulate different usage behaviors in real-world scenarios. 
Such behavior highlights the need of developing models that can adapt to both battery aging and varying load conditions. In the next experimental section, our proposed SeqBattNet will demonstrate its ability to meet these requirements.

\section{Experiment}\label{sec:experiments}

\subsection{Experimental Setup}
Our proposed SeqBattNet is developed using the PyTorch framework. AdamW optimizer~\cite{llugsi2021comparison} is used to train the model. All experiments are performed on an NVIDIA 4080 GPU with 16GB of memory. Each experiment is trained for 1000 epochs with a batch size of 128. We use PyTorch’s \texttt{ReduceLROnPlateau}~\cite{paszke2019pytorch} with an initial learning rate of $2\times10^{-3}$ (mode=min). If the validation loss does not improve for 5 epochs, the learning rate is reduced by a factor of 0.8, with a lower bound of $1\times10^{-4}$. To minimize variability in the training process, each experiment is repeated 5 times, and the final prediction is calculated as the average of these multiple runs.



\subsection{Effect of Initial Sequence Length }
\begin{figure}[H]
    \centering
    \includegraphics[width=1\textwidth]{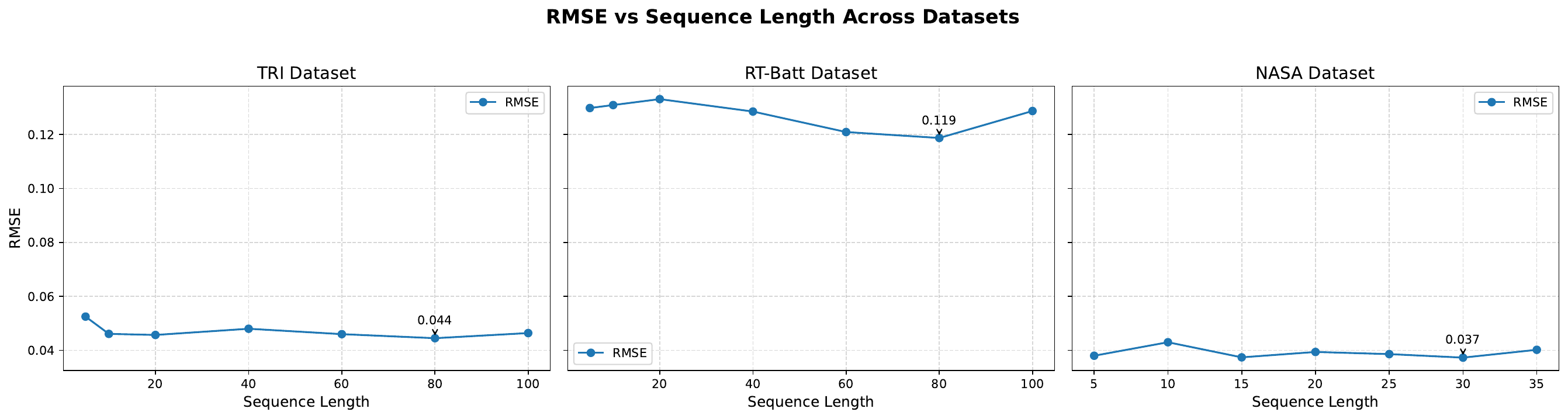}  
    \caption{Impact of the initial sequence length used for aging adaptation on TRI, RT-Batt, NASA dataset.}
    \label{fig:Sequence_Length_vs_RMSE}
\end{figure}

The length of the initial sequence data $\mathbf{x}$ in each discharge cycle, used for aging adaptation, affects the prediction of aging adaptation parameters. We do not evaluate excessively long sequence lengths for aging adaptation within each cycle, since in practice users need to monitor battery states as early as possible. As shown in Figure~\ref{fig:Sequence_Length_vs_RMSE}, the TRI and RT-Batt datasets achieve their best performance with a sequence length of 80, while the NASA dataset performs best with a sequence length of 30. One of the reasons the TRI and RT-Batt datasets require a common initial segment length is that they use the same battery cell type A123 APR18650M1A. Each battery type, however, requires a different effective initial discharge length for aging detection. These sequence lengths are employed in the subsequent experiments.

\subsection{Effect of the Encoder on the Aging Adaptation task}
\begin{figure}[H]
    \centering
    \includegraphics[width=1\textwidth]{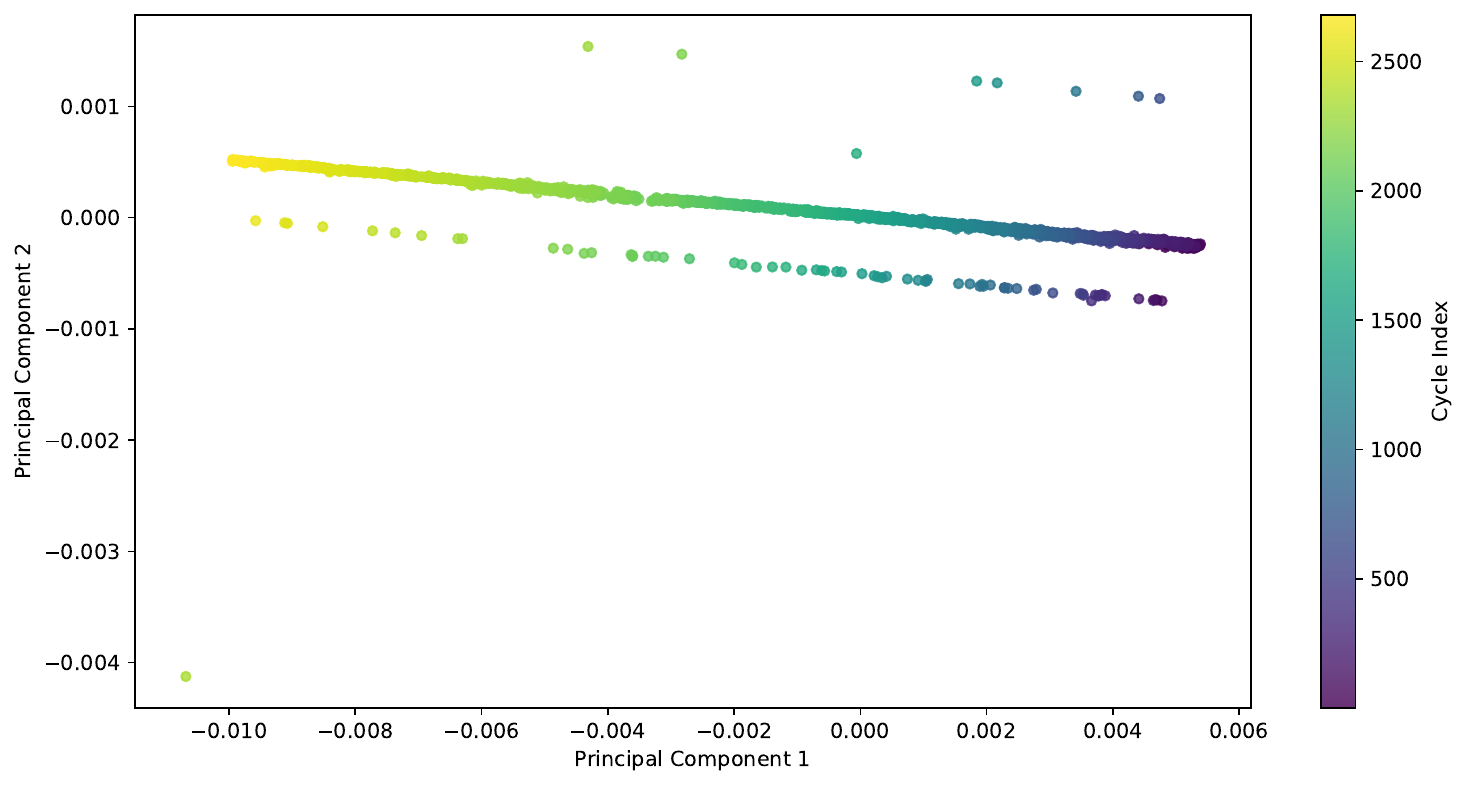}  
    \caption{Aging trajectories visualized using PCA on the Encoder embeddings for Cell 1-2 of the RT-Batt Dataset.}
    \label{fig:Aging_Trajectory}
\end{figure}

Figure \ref{fig:Aging_Trajectory} illustrates the effect of the encoder on aging adaptation by projecting the learned embeddings, obtained from the encoder’s final layer, onto two principal components using principal component analysis (PCA)~\cite{abdi2010principal}. The trajectories show a clear progression with respect to the cycle index, where earlier cycles (purple/blue) gradually diverge from later cycles (green/yellow). This smooth transition indicates that the encoder effectively captures the temporal evolution of battery degradation. The clustering and gradual shifts across principal components suggest that our model is sensitive to aging dynamics, enabling it to distinguish between early-life and late-life behaviors.

\subsection{Loss Function Validation}
Table~\ref{tab:loss_function} compares our proposed loss function with the standard MSE~\cite{chen2022novel} and the baseline loss function from \cite{wang2023inherently}, on which our method is built. Overall, the MSE yields the poorest results, with RMSE values as high as 0.1339 on TRI-b1c2 and 0.0965 on RT-Batt-1-3. The baseline loss function performs better, reducing RMSE to 0.0624 on TRI-b1c4 and 0.0579 on RT-Batt-1-2. In contrast, our proposed loss function consistently achieves the best results across all datasets, e.g., lowering RMSE to 0.028 on TRI-b1c4 and 0.0346 on RT-Batt-1-2, along with corresponding MAE and MAPE improvements. On the NASA dataset, our loss also yields the lowest errors (e.g., RMSE of 0.0253 on RW5), outperforming both MSE (0.0329) and the baseline (0.0301). These gains demonstrate that the weighting scheme—designed to emphasize errors toward the end of the prediction sequence—helps the model capture long-term dependencies more effectively, thereby enhancing prediction accuracy.

\begin{table}[H]
\centering
\caption{Comparison of the proposed loss function with baseline methods.}
\resizebox{1\textwidth}{!}{%
\begin{tabular}{cccccccccccccc}
\toprule
 & & \multicolumn{3}{c}{\textbf{MSE (1.2)}} & \multicolumn{3}{c}{\cite{wang2023inherently} (1.4)} & \multicolumn{3}{c}{\textbf{Ours} (0.9)} \\

\cmidrule(lr){3-5} \cmidrule(lr){6-8} \cmidrule(lr){9-11} 
\textbf{Dataset} & \textbf{Battery} & \textbf{RMSE} & 
\textbf{MAE} & \textbf{MAPE (\%)} & \textbf{RMSE} & \textbf{MAE} & \textbf{MAPE (\%)} & \textbf{RMSE} & \textbf{MAE} & \textbf{MAPE (\%)} \\
\midrule
\multirow{3}{*}{TRI}  
  & b1c4  & 0.1299 & 0.1003 & 0.0345 & 0.0624 & 0.0252 & 0.0096 & 0.028   & 0.0113  & 0.0042  \\
  & b1c2  & 0.1339 & 0.1038 & 0.0355 & 0.068  & 0.0261 & 0.0099 & 0.0416  & 0.0168  & 0.0063 \\
\midrule
\multirow{3}{*}{RT-Batt} 
    & 1-2 & 0.0902 & 0.0445 & 0.0155 & 0.0579 & 0.0196 & 0.0073 & 0.0346 & 0.012  & 0.0044 \\
    & 1-3 & 0.0965 & 0.0499 & 0.0172 & 0.0593 & 0.0199 & 0.0074 & 0.0491 & 0.0171 & 0.0063 \\
\midrule
\multirow{3}{*}{NASA} 
   & RW4 & 0.0319 & 0.0226 & 0.0064 & 0.0288 & 0.0194 & 0.0056 & 0.0257  & 0.0191  & 0.0055 \\
   & RW5 & 0.0329 & 0.0238 & 0.0067 & 0.0301 & 0.0204 & 0.0059 & 0.0253  & 0.0191  & 0.0055 \\
\bottomrule
\end{tabular}
} 
\label{tab:loss_function}
\end{table}

\subsection{Voltage Prediction Validation}
Table~\ref{tab:01} summarizes the voltage prediction results across three datasets and multiple model families. 
(1) Classical sequence models (LSTM, GRU, Transformer) improve significantly over BattNN~\cite{wang2023inherently}, reducing RMSE by more than $80\%$ on TRI (from $\approx 0.58$ to $\approx 0.07$) and by $75\%$ on RT-Batt (from $\approx 0.55$ to $\approx 0.13$). 
(2) Variants of SeqBattNet are obtained by replacing the encoder with different networks. Specifically, HRM-LSTM, HRM-GRU, and HRM-Transformer correspond to using LSTM, GRU, and Transformer encoders within the HRM framework. 
(3) Our proposed SeqBattNet with an HRM-GRU encoder consistently achieves the best performance on the TRI dataset (RMSE $\approx 0.028$ for b1c4) and the RT-Batt dataset (RMSE $\approx 0.035$ for 1-2), representing a $50$--$60\%$ error reduction compared to other SeqBattNet variants. 
(4) On the NASA dataset, the HRM-GRU encoder ranks second-best (RMSE $\approx 0.025$--$0.026$), with the FNN encoder achieving the best results (RMSE $\approx 0.024$), indicating dataset-specific encoder suitability. 
(5) In terms of computational efficiency, SeqBattNet achieves competitive runtimes of $\approx 30$--$90$ seconds per experiment, which is comparable to or faster than classical sequence models while delivering substantially lower errors. 
Overall, these results demonstrate that our proposed SeqBattNet with HRM-GRU generalizes well across different working conditions: identical discharge load profiles across cells and cycles (TRI dataset), different profiles across cells but consistent across cycles (RT-Batt dataset), and varying profiles across both cells and cycles (NASA dataset).

\begin{table}[H]
\centering
\caption{Performance comparison of our proposed voltage prediction model and other approaches.}
\resizebox{\textwidth}{!}{%
\begin{tabular}{llcccccccccccccccc}
\toprule[2pt]
& & \multicolumn{4}{c}{\textbf{BattNN~\cite{wang2023inherently} (0.0)}} 
  & \multicolumn{4}{c}{\textbf{LSTM (0.1)}} 
  & \multicolumn{4}{c}{\textbf{GRU (0.2)}} 
  & \multicolumn{4}{c}{\textbf{Transformer (0.3)}} \\
\cmidrule(lr){3-6} \cmidrule(lr){7-10} \cmidrule(lr){11-14} \cmidrule(lr){15-18}
\textbf{Dataset} & \textbf{Battery} & \textbf{RMSE} & \textbf{MAE} & \textbf{MAPE (\%)} & \textbf{Time (s)} & \textbf{RMSE} & \textbf{MAE} & \textbf{MAPE (\%)} & \textbf{Time (s)} & \textbf{RMSE} & \textbf{MAE} & \textbf{MAPE (\%)} & \textbf{Time (s)} & \textbf{RMSE} & \textbf{MAE} & \textbf{MAPE (\%)} & \textbf{Time (s)} \\
\midrule
\multirow{3}{*}{TRI}  
    & b1c4  & 0.5781  & 0.5378  & 0.1765  & 2.4481  & 0.0712  & 0.034   & 0.0127  & 10.0834  & 0.0713  & 0.034   & 0.0127  & 9.4537  & 0.0762 & 0.0357 & 0.0133 & 32.6908 \\
    & b1c2  & 0.5827  & 0.5422  & 0.1777  & 2.5445  & 0.0707  & 0.0327  & 0.0121  & 12.3596  & 0.0704  & 0.0314  & 0.0117  &12.2257  & 0.0744 & 0.0323 & 0.0121 & 41.6712 \\
\midrule
\multirow{3}{*}{RT-Batt} 
    & 1-2   & 0.5536  & 0.5045  & 0.1619 & 137.8269  & 0.1303  & 0.0516  & 0.0189  & 9.5352  & 0.2197  & 0.1027  & 0.0353  & 10.1433  & 0.145 & 0.0699 & 0.025 & 20.3147 \\
    & 1-3   & 0.5294  & 0.481   & 0.1546 & 95.7069  & 0.134   & 0.0518  & 0.0191  & 6.3358  & 0.1114  & 0.0498  & 0.0177  & 6.3457  & 0.1513 & 0.0741 & 0.0265 & 12.944\\
\midrule
\multirow{3}{*}{NASA} 
    & RW4   & 0.0502  &  0.0397 & 0.0112  & 1.7414  & 0.0297 & 0.0208 & 0.0059  & 0.1441  & 0.0272  & 0.0203  & 0.0058  & 0.1446  & 0.0251 & 0.017 & 0.0049 & 0.1884\\
    & RW5   & 0.0521  & 0.0406  & 0.0115  & 1.8687  & 0.0308 & 0.022  & 0.0062  & 0.1712  & 0.0281  & 0.021   & 0.0061  & 0.1618  & 0.027 & 0.0181 & 0.0052 & 0.1966\\
\midrule\midrule

& & \multicolumn{4}{c}{\textbf{SeqBattNet (0.4)}}
  & \multicolumn{4}{c}{\textbf{SeqBattNet (0.5)}} 
  & \multicolumn{4}{c}{\textbf{SeqBattNet (0.6)}} 
  & \multicolumn{4}{c}{\textbf{SeqBattNet (0.7)}} \\
& & \multicolumn{4}{c}{LSTM for Encoder} 
    & \multicolumn{4}{c}{GRU for Encoder} 
    & \multicolumn{4}{c}{Transformer for Encoder} 
    & \multicolumn{4}{c}{FNN for Encoder} \\
\cmidrule(lr){3-6} \cmidrule(lr){7-10} \cmidrule(lr){11-14} \cmidrule(lr){15-18}
\textbf{Dataset} & \textbf{Battery} & \textbf{RMSE} & \textbf{MAE} & \textbf{MAPE (\%)} & \textbf{Time (s)} & \textbf{RMSE} & \textbf{MAE} & \textbf{MAPE (\%)} & \textbf{Time (s)} & \textbf{RMSE} & \textbf{MAE} & \textbf{MAPE (\%)} & \textbf{Time (s)} & \textbf{RMSE} & \textbf{MAE} & \textbf{MAPE (\%)} & \textbf{Time (s)} \\
\midrule
\multirow{3}{*}{TRI}  
    & b1c4  & 0.0684  & 0.0284  & 0.0108  & 60.817   & 0.0742 & 0.0319 & 0.012   & 59.5293 & 0.0483 & 0.0211 & 0.0079 & 55.9238 & 0.0647  & 0.0268  & 0.0101  & 57.6439   \\
    & b1c2  & 0.0677  & 0.028   & 0.0105  & 78.6893  & 0.0736 & 0.0304 & 0.0114  & 76.7242 & 0.0528 & 0.0223 & 0.0083 & 73.3107 & 0.0653  &  0.0262  & 0.0099  & 73.64     \\
\midrule
\multirow{3}{*}{RT-Batt} 
    & 1-2   & 0.0827  & 0.032   & 0.0117  & 58.8327  & 0.0827 & 0.0346 & 0.0124  & 55.5968 & 0.0824  & 0.0323  & 0.0118  & 56.9902 & 0.0826  & 0.0319  & 0.0117  &  51.1914   \\
    & 1-3   & 0.0723  & 0.0283  & 0.0103  & 38.4323  & 0.0845 & 0.0393 & 0.0139  & 36.5965 & 0.0726  & 0.0296  & 0.0108  & 37.1704 & 0.0721  & 0.0284  & 0.0104  &  33.3752   \\
\midrule
\multirow{3}{*}{NASA} 
    & RW4   & 0.0259  & 0.0175  & 0.005  & 0.6694  & 0.0343 & 0.0234 & 0.0066  & 0.6745 &  0.0274 & 0.0181 & 0.0051  & 0.6676  & 0.0239  & 0.0155  & 0.0044  & 0.5531   \\
    & RW5   & 0.0264  & 0.0177  & 0.005  & 0.7327  & 0.0359 & 0.0249 & 0.0071  & 0.6741 & 0.0279 & 0.0183 & 0.0052  & 0.6768  & 0.0248  & 0.0159  & 0.0046  & 0.609   \\

\midrule\midrule

& & \multicolumn{4}{c}{\textbf{SeqBattNet (0.8)}}
  & \multicolumn{4}{c}{\textbf{SeqBattNet (0.9)}} 
  & \multicolumn{4}{c}{\textbf{SeqBattNet (1.0)}} 
  & \multicolumn{4}{c}{\textbf{SeqBattNet (1.1)}} \\
& & \multicolumn{4}{c}{HRM-LSTM for Encoder} 
    & \multicolumn{4}{c}{HRM-GRU for Encoder \textit{(Proposed)}} 
    & \multicolumn{4}{c}{HRM-Transformer for Encoder} 
    & \multicolumn{4}{c}{FNN for Encoder} \\
\cmidrule(lr){3-6} \cmidrule(lr){7-10} \cmidrule(lr){11-14} \cmidrule(lr){15-18}
\textbf{Dataset} & \textbf{Battery} & \textbf{RMSE} & \textbf{MAE} & \textbf{MAPE (\%)} & \textbf{Time (s)} & \textbf{RMSE} & \textbf{MAE} & \textbf{MAPE (\%)} & \textbf{Time (s)} & \textbf{RMSE} & \textbf{MAE} & \textbf{MAPE (\%)} & \textbf{Time (s)} & \textbf{RMSE} & \textbf{MAE} & \textbf{MAPE (\%)} & \textbf{Time (s)} \\
\midrule
\multirow{3}{*}{TRI}  
    & b1c4  & 0.0637  & 0.0255  & 0.0097  & 68.6676  &  0.028   & 0.0113  & 0.0042 & 68.819  & 0.0701 & 0.0298 & 0.0113 & 109.5333 & 0.0712  & 0.0317  & 0.0118  & 68.2895    \\
    & b1c2  & 0.0676  & 0.0265  & 0.0101  & 88.3893  & 0.0416  & 0.0168  & 0.0063 & 85.4199 & 0.0692 & 0.0291 & 0.0109 & 138.7852 & 0.0712  & 0.032  & 0.0118  & 87.8615    \\
\midrule
\multirow{3}{*}{RT-Batt} 
    & 1-2   & 0.0727  & 0.0264  & 0.0097  & 68.6494  & 0.0346 & 0.012  & 0.0044 & 70.2151 & 0.0827 & 0.032  & 0.0117 & 129.469 &  0.0813 & 0.0318  & 0.0116  &  72.7671   \\
    & 1-3   & 0.0683  & 0.0324  & 0.0115  & 46.1185  & 0.0491 & 0.0171 & 0.0063 & 46.2879 & 0.0723 & 0.0286 & 0.0105 & 91.5932 &  0.0719 & 0.029  & 0.0106  & 46.3756    \\
\midrule
\multirow{3}{*}{NASA} 
    & RW4   & 0.0307  & 0.0209  & 0.006   & 0.8571  & 0.0257  & 0.0191  & 0.0055 & 0.8675 & 0.029  & 0.0194 & 0.0055 & 1.7621 & 0.0315  & 0.0227  & 0.0065  &  0.87   \\
    & RW5   & 0.0313  & 0.0212  & 0.0061  & 0.9251  & 0.0253  & 0.0191  & 0.0055 & 0.8907 & 0.0293 & 0.0195 & 0.0055 & 1.7676 & 0.0325  & 0.0237  & 0.0068  &  0.9289   \\
\bottomrule[2pt]
\end{tabular}
}
\label{tab:01}
\end{table}

\begin{figure}[H]
    \centering
    \begin{subfigure}[b]{0.32\textwidth}
        \includegraphics[width=\textwidth]{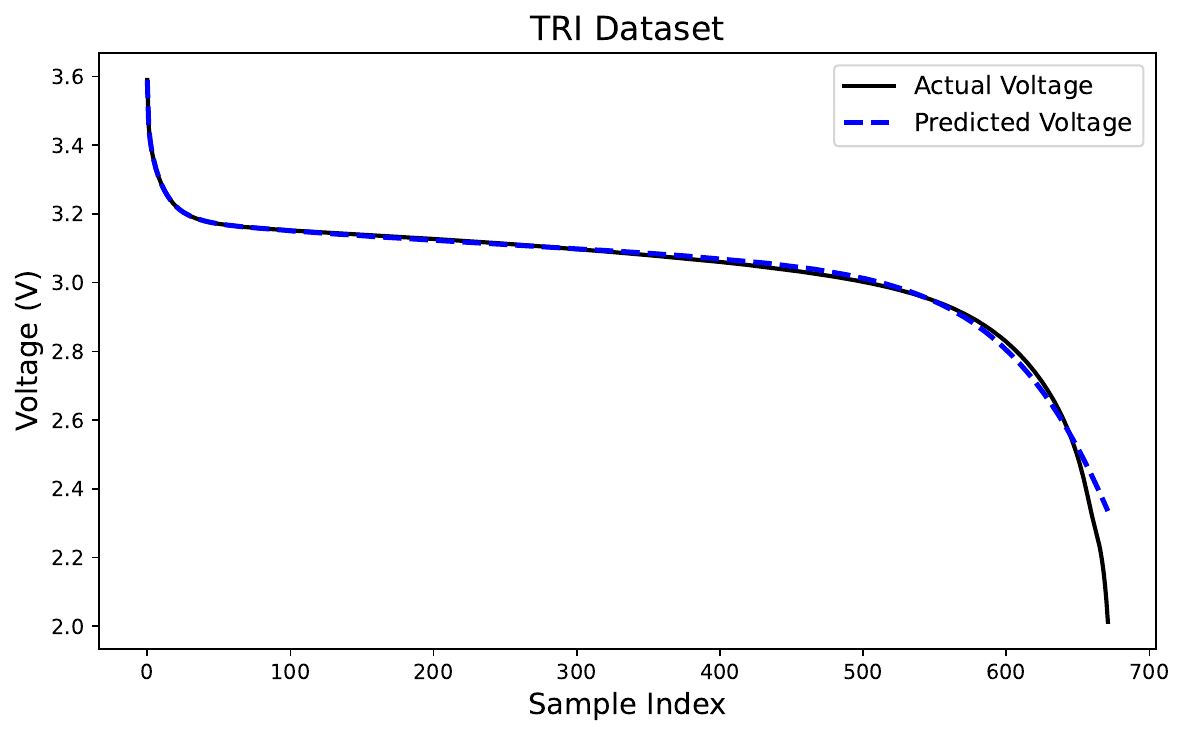}
        \label{fig:TRI_visualization}
    \end{subfigure}
    \hfill
    \begin{subfigure}[b]{0.32\textwidth}
        \includegraphics[width=\textwidth]{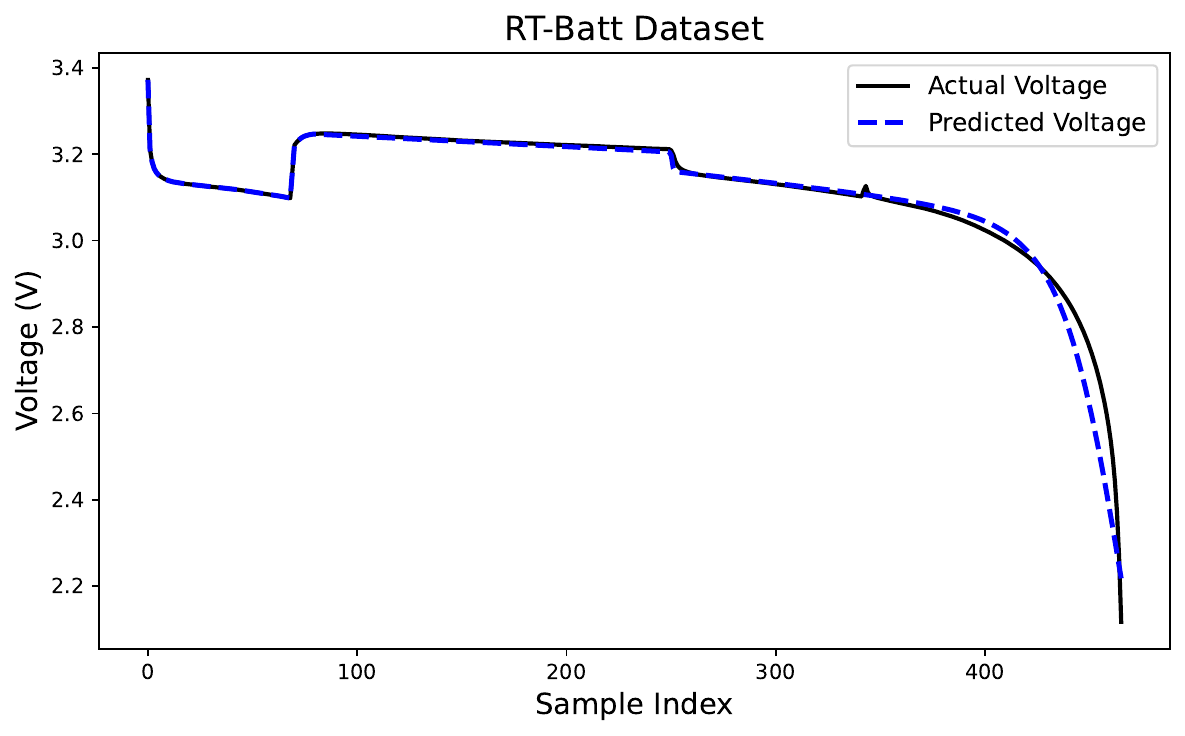}
        \label{fig:RT_Batt_visualization}
    \end{subfigure}
    \hfill
    \begin{subfigure}[b]{0.32\textwidth}
        \includegraphics[width=\textwidth]{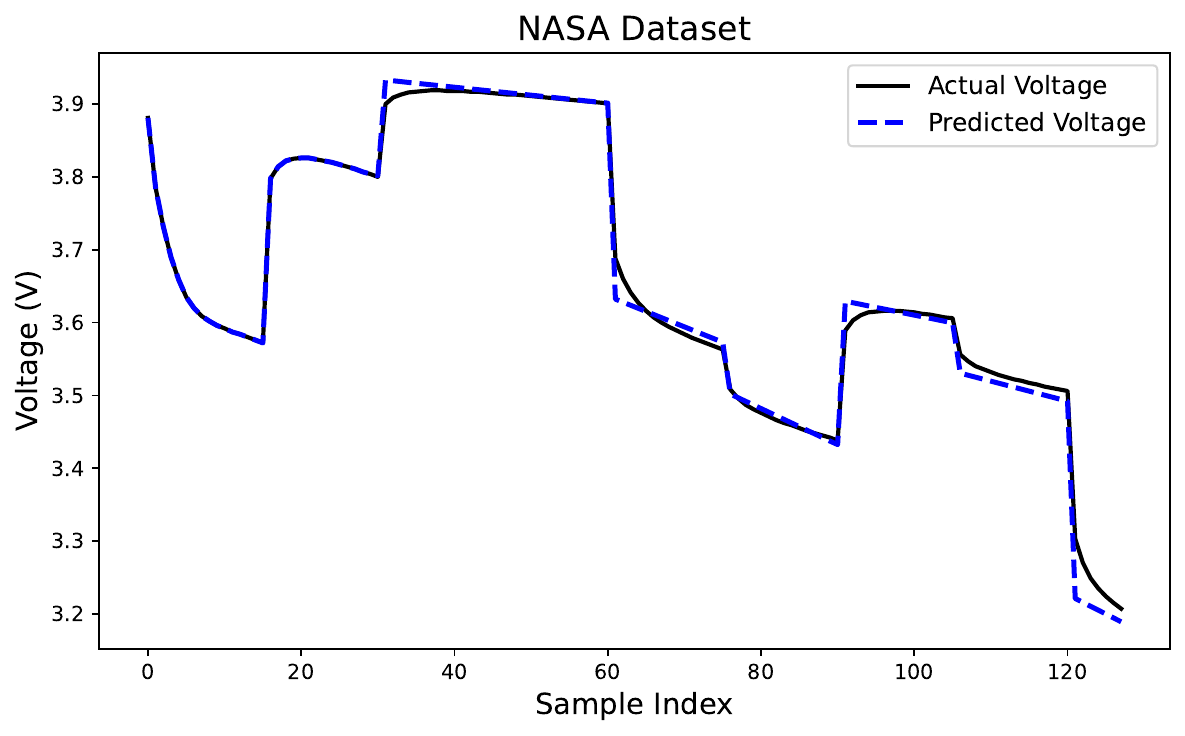}
        \label{fig:NASA_visualization}
    \end{subfigure}
    \caption{Visualization of terminal voltage prediction using our proposed SeqBattNet on the TRI, RT-Batt, and NASA datasets.}
    \label{fig:three_images}
\end{figure}

Figure~\ref{fig:three_images} illustrates the performance of our proposed SeqBattNet in predicting terminal voltage across three datasets: TRI, RT-Batt, and NASA. 
In each case, the predicted voltage trajectories (blue dashed lines) closely follow the actual measured voltages (black solid lines), demonstrating the model’s ability to capture both the overall discharge trend and local fluctuations. For the TRI dataset, the model accurately tracks the smooth discharge curve and the steep voltage drop near the end-of-discharge. For the RT-Batt dataset, SeqBattNet successfully captures the irregular step-like voltage behaviors caused by varying load conditions. For the NASA dataset, the model adapts to highly dynamic discharge profiles with sharp transitions and variable plateau regions. These results show that SeqBattNet generalizes effectively across different operating conditions, yielding robust and reliable voltage predictions. 

\subsection{Discussion and Limitations}
In this study, we focus solely on voltage and current as model inputs, while excluding the influence of ambient and battery temperature. The datasets employed are collected under controlled laboratory conditions, where temperature variations are minimal and do not fully reflect real-world operating environments. However, in practical applications, temperature is a critical factor that significantly affects battery dynamics, including internal resistance, capacity fade, and voltage response. Consequently, our future work will extend the model by incorporating temperature features (both ambient and cell temperature) as additional inputs. Moreover, we aim to validate the model on real-world datasets that capture diverse environmental and operational conditions, thereby ensuring its robustness and generalizability beyond laboratory settings.

\section{Conclusion}\label{sec:conclusion}
This work highlights the potential of \textbf{SeqBattNet}, a discrete-state physics-informed neural network with aging adaptation for battery modeling in the discharge process. SeqBattNet consists of two main components: (i) an encoder that produces cycle-specific adaptation parameters, and (ii) a decoder that leverages these parameters together with the input current to predict the terminal voltage. The proposed model demonstrates outstanding performance across multiple benchmark datasets, outperforming classical sequence models and baselines.  In future work, we plan to extend SeqBattNet by incorporating temperature features and validating it on real-world datasets, aiming to further improve robustness and practical applicability in battery management systems.

\section*{Contact Information}
For access to the code and further information about this proposed system, please contact AIWARE Limited Company at: \url{https://aiware.website/Contact}

\bibliographystyle{plain}
\bibliography{cas-refs}

\begin{thebibliography}{10}

\bibitem{abdi2010principal}
Herv{\'e} Abdi and Lynne~J Williams.
\newblock Principal component analysis.
\newblock {\em Wiley interdisciplinary reviews: computational statistics}, 2(4):433--459, 2010.

\bibitem{ali2024comparison}
Haider Adel~Ali Ali, Luc~HJ Raijmakers, Kudakwashe Chayambuka, Dmitri~L Danilov, Peter~HL Notten, and R{\"u}diger-A Eichel.
\newblock A comparison between physics-based li-ion battery models.
\newblock {\em Electrochimica Acta}, 493:144360, 2024.

\bibitem{ba2016layer}
Jimmy~Lei Ba, Jamie~Ryan Kiros, and Geoffrey~E Hinton.
\newblock Layer normalization.
\newblock {\em arXiv preprint arXiv:1607.06450}, 2016.

\bibitem{biggio2023ageing}
Luca Biggio, Tommaso Bendinelli, Chetan Kulkarni, and Olga Fink.
\newblock Ageing-aware battery discharge prediction with deep learning.
\newblock {\em Applied Energy}, 346:121229, 2023.

\bibitem{bole2014adaptation}
Brian Bole, Chetan~S Kulkarni, and Matthew Daigle.
\newblock Adaptation of an electrochemistry-based li-ion battery model to account for deterioration observed under randomized use.
\newblock In {\em Annual conference of the PHM society}, volume~6, 2014.

\bibitem{cen2020lithium}
Zhaohui Cen and Pierre Kubiak.
\newblock Lithium-ion battery soc/soh adaptive estimation via simplified single particle model.
\newblock {\em International Journal of Energy Research}, 44(15):12444--12459, 2020.

\bibitem{chen2022novel}
Xi~Chen, Ruyi Yu, Sajid Ullah, Dianming Wu, Zhiqiang Li, Qingli Li, Honggang Qi, Jihui Liu, Min Liu, and Yundong Zhang.
\newblock A novel loss function of deep learning in wind speed forecasting.
\newblock {\em Energy}, 238:121808, 2022.

\bibitem{chen2019particle}
Zonghai Chen, Han Sun, Guangzhong Dong, Jingwen Wei, and JI~Wu.
\newblock Particle filter-based state-of-charge estimation and remaining-dischargeable-time prediction method for lithium-ion batteries.
\newblock {\em Journal of Power Sources}, 414:158--166, 2019.

\bibitem{csomos2023comparison}
Bence Csom{\'o}s, Szabolcs Kocsis~Sz{\"u}rke, and D{\'e}nes Fodor.
\newblock Comparison of coupled electrochemical and thermal modelling strategies of 18650 li-ion batteries in finite element analysis—a review.
\newblock {\em Materials}, 16(24):7613, 2023.

\bibitem{dey2017gate}
Rahul Dey and Fathi~M Salem.
\newblock Gate-variants of gated recurrent unit (gru) neural networks.
\newblock In {\em 2017 IEEE 60th international midwest symposium on circuits and systems (MWSCAS)}, pages 1597--1600. IEEE, 2017.

\bibitem{gabbar2021review}
Hossam~A Gabbar, Ahmed~M Othman, and Muhammad~R Abdussami.
\newblock Review of battery management systems (bms) development and industrial standards.
\newblock {\em Technologies}, 9(2):28, 2021.

\bibitem{girshick2015fast}
Ross Girshick.
\newblock Fast r-cnn.
\newblock In {\em Proceedings of the IEEE international conference on computer vision}, pages 1440--1448, 2015.

\bibitem{gu2000thermal}
Wa~B Gu and Chaoyang~Y Wang.
\newblock Thermal-electrochemical modeling of battery systems.
\newblock {\em Journal of The Electrochemical Society}, 147(8):2910, 2000.

\bibitem{hatherall2023remaining}
Ollie Hatherall, Mona~Faraji Niri, Anup Barai, Yi~Li, and James Marco.
\newblock Remaining discharge energy estimation for lithium-ion batteries using pattern recognition and power prediction.
\newblock {\em Journal of Energy Storage}, 64:107091, 2023.

\bibitem{hochreiter1997long}
Sepp Hochreiter and J{\"u}rgen Schmidhuber.
\newblock Long short-term memory.
\newblock {\em Neural computation}, 9(8):1735--1780, 1997.

\bibitem{lai2022remaining}
Xin Lai, Yunfeng Huang, Huanghui Gu, Xuebing Han, Xuning Feng, Haifeng Dai, Yuejiu Zheng, and Minggao Ouyang.
\newblock Remaining discharge energy estimation for lithium-ion batteries based on future load prediction considering temperature and ageing effects.
\newblock {\em Energy}, 238:121754, 2022.

\bibitem{lai2018comparative}
Xin Lai, Yuejiu Zheng, and Tao Sun.
\newblock A comparative study of different equivalent circuit models for estimating state-of-charge of lithium-ion batteries.
\newblock {\em Electrochimica Acta}, 259:566--577, 2018.

\bibitem{liang2024hybrid}
Junyuan Liang, Hui Liu, and Ning-Cong Xiao.
\newblock A hybrid approach based on deep neural network and double exponential model for remaining useful life prediction.
\newblock {\em Expert Systems with Applications}, 249:123563, 2024.

\bibitem{liashchynskyi2019grid}
Petro Liashchynskyi and Pavlo Liashchynskyi.
\newblock Grid search, random search, genetic algorithm: a big comparison for nas.
\newblock {\em arXiv preprint arXiv:1912.06059}, 2019.

\bibitem{liaw2004modeling}
Bor~Yann Liaw, Ganesan Nagasubramanian, Rudolph~G Jungst, and Daniel~H Doughty.
\newblock Modeling of lithium ion cells—a simple equivalent-circuit model approach.
\newblock {\em Solid state ionics}, 175(1-4):835--839, 2004.

\bibitem{liu2020novel}
Shiqi Liu, Junhua Wang, Qisheng Liu, Jia Tang, Haolu Liu, Yang Zhou, and Xingya Pan.
\newblock A novel discharge mode identification method for series-connected battery pack online state-of-charge estimation over a wide life scale.
\newblock {\em IEEE Transactions on Power Electronics}, 36(1):326--341, 2020.

\bibitem{liu2024hybrid}
Yunpeng Liu, Bo~Hou, Moin Ahmed, Zhiyu Mao, Jiangtao Feng, and Zhongwei Chen.
\newblock A hybrid deep learning approach for remaining useful life prediction of lithium-ion batteries based on discharging fragments.
\newblock {\em Applied Energy}, 358:122555, 2024.

\bibitem{llugsi2021comparison}
Ricardo Llugsi, Samira El~Yacoubi, Allyx Fontaine, and Pablo Lupera.
\newblock Comparison between adam, adamax and adam w optimizers to implement a weather forecast based on neural networks for the andean city of quito.
\newblock In {\em 2021 IEEE Fifth Ecuador Technical Chapters Meeting (ETCM)}, pages 1--6. IEEE, 2021.

\bibitem{ma2023two}
Guijun Ma, Zidong Wang, Weibo Liu, Jingzhong Fang, Yong Zhang, Han Ding, and Ye~Yuan.
\newblock A two-stage integrated method for early prediction of remaining useful life of lithium-ion batteries.
\newblock {\em Knowledge-Based Systems}, 259:110012, 2023.

\bibitem{ma2022real}
Guijun Ma, Songpei Xu, Benben Jiang, Cheng Cheng, Xin Yang, Yue Shen, Tao Yang, Yunhui Huang, Han Ding, and Ye~Yuan.
\newblock Real-time personalized health status prediction of lithium-ion batteries using deep transfer learning.
\newblock {\em Energy \& Environmental Science}, 15(10):4083--4094, 2022.

\bibitem{ma2024accurate}
Liang Ma, Jinpeng Tian, Tieling Zhang, Qinghua Guo, and Chunsheng Hu.
\newblock Accurate and efficient remaining useful life prediction of batteries enabled by physics-informed machine learning.
\newblock {\em Journal of Energy Chemistry}, 91:512--521, 2024.

\bibitem{noel2019fear}
Lance Noel, Gerardo~Zarazua De~Rubens, Benjamin~K Sovacool, and Johannes Kester.
\newblock Fear and loathing of electric vehicles: The reactionary rhetoric of range anxiety.
\newblock {\em Energy research \& social science}, 48:96--107, 2019.

\bibitem{nosouhian2021review}
Shiva Nosouhian, Fereshteh Nosouhian, and Abbas~Kazemi Khoshouei.
\newblock A review of recurrent neural network architecture for sequence learning: Comparison between lstm and gru.
\newblock 2021.

\bibitem{paszke2019pytorch}
Adam Paszke, Sam Gross, Francisco Massa, Adam Lerer, James Bradbury, Gregory Chillemi, Luca Antiga, Alban Desmaison, Andreas Tejani, Soumith Chanan, et~al.
\newblock Pytorch: An imperative style, high-performance deep learning library.
\newblock {\em Advances in Neural Information Processing Systems}, 32, 2019.

\bibitem{qu2019neural}
Jiantao Qu, Feng Liu, Yuxiang Ma, and Jiaming Fan.
\newblock A neural-network-based method for rul prediction and soh monitoring of lithium-ion battery.
\newblock {\em IEEE access}, 7:87178--87191, 2019.

\bibitem{quinones2018remaining}
Facundo Qui{\~n}ones, RH~Milocco, and SG~Real.
\newblock Remaining discharge-time prediction for batteries using the lambert function.
\newblock {\em Journal of Power Sources}, 400:256--263, 2018.

\bibitem{sakile2022estimation}
Rajakumar Sakile and Umesh~Kumar Sinha.
\newblock Estimation of lithium-ion battery state of charge for electric vehicles using a nonlinear state observer.
\newblock {\em Energy Storage}, 4(2):e290, 2022.

\bibitem{severson2019data}
Kristen~A Severson, Peter~M Attia, Norman Jin, Nicholas Perkins, Benben Jiang, Zi~Yang, Michael~H Chen, Muratahan Aykol, Patrick~K Herring, Dimitrios Fraggedakis, et~al.
\newblock Data-driven prediction of battery cycle life before capacity degradation.
\newblock {\em Nature Energy}, 4(5):383--391, 2019.

\bibitem{somakettarin2019characterization}
Natthawuth Somakettarin and Achara Pichetjamroen.
\newblock Characterization of a practical-based ohmic series resistance model under life-cycle changes for a lithium-ion battery.
\newblock {\em Energies}, 12(20):3888, 2019.

\bibitem{tan2025batterylife}
Ruifeng Tan, Weixiang Hong, Jiayue Tang, Xibin Lu, Ruijun Ma, Xiang Zheng, Jia Li, Jiaqiang Huang, and Tong-Yi Zhang.
\newblock Batterylife: A comprehensive dataset and benchmark for battery life prediction.
\newblock In {\em Proceedings of the 31st ACM SIGKDD Conference on Knowledge Discovery and Data Mining V. 2}, pages 5789--5800, 2025.

\bibitem{tran2021comparative}
Manh-Kien Tran, Andre DaCosta, Anosh Mevawalla, Satyam Panchal, and Michael Fowler.
\newblock Comparative study of equivalent circuit models performance in four common lithium-ion batteries: Lfp, nmc, lmo, nca.
\newblock {\em Batteries}, 7(3):51, 2021.

\bibitem{tu2024remaining}
Hao Tu, Manashita Borah, Scott Moura, Yebin Wang, and Huazhen Fang.
\newblock Remaining discharge energy prediction for lithium-ion batteries over broad current ranges: A machine learning approach.
\newblock {\em Applied Energy}, 376:124086, 2024.

\bibitem{vaswani2017attention}
Ashish Vaswani, Noam Shazeer, Niki Parmar, Jakob Uszkoreit, Llion Jones, Aidan~N Gomez, {\L}ukasz Kaiser, and Illia Polosukhin.
\newblock Attention is all you need.
\newblock {\em Advances in neural information processing systems}, 30, 2017.

\bibitem{wang2023inherently}
Fujin Wang, Quanquan Zhi, Zhibin Zhao, Zhi Zhai, Yingkai Liu, Huan Xi, Shibin Wang, and Xuefeng Chen.
\newblock Inherently interpretable physics-informed neural network for battery modeling and prognosis.
\newblock {\em IEEE Transactions on Neural Networks and Learning Systems}, 2023.

\bibitem{wang2025hierarchical}
Guan Wang, Jin Li, Yuhao Sun, Xing Chen, Changling Liu, Yue Wu, Meng Lu, Sen Song, and Yasin~Abbasi Yadkori.
\newblock Hierarchical reasoning model.
\newblock {\em arXiv preprint arXiv:2506.21734}, 2025.

\bibitem{wang2025adaptive}
Shuo Wang, Yonghong Xu, Hongguang Zhang, Rao Kuang, Jian Zhang, Baicheng Liu, Fubin Yang, and Yujie Zhang.
\newblock An adaptive cubature kalman filter algorithm based on singular value decomposition for joint estimation of state of charge and state of power for lithium-ion batteries under wide temperature range.
\newblock {\em Ionics}, 31(1):345--365, 2025.

\bibitem{wang2020framework}
Yujie Wang and Zonghai Chen.
\newblock A framework for state-of-charge and remaining discharge time prediction using unscented particle filter.
\newblock {\em Applied Energy}, 260:114324, 2020.

\end{thebibliography}
\end{document}